\documentclass[12pt]{article}
\usepackage{newtxtext,newtxmath}
\usepackage{graphicx}
\usepackage[letterpaper,margin=1in]{geometry}
\usepackage{multirow}%
\usepackage{amsmath,amsfonts}%
\usepackage[title]{appendix}%
\usepackage{xcolor}%
\usepackage{textcomp}%
\usepackage{manyfoot}%
\usepackage{booktabs}%
\usepackage{algorithm}%
\usepackage{algorithmicx}%
\usepackage{algpseudocode}%
\usepackage{listings}%

\usepackage{hyperref}       
\usepackage{url}            
\usepackage{nicefrac}       
\usepackage{microtype}      
\usepackage{lipsum}
\usepackage{comment}
\usepackage{caption}
\usepackage{subcaption}
\usepackage{multicol} %
\usepackage[inline]{enumitem}
\usepackage{tcolorbox}
\usepackage{framed} %
\usepackage{makecell}
\usepackage{hhline}
\usepackage{framed}
\usepackage[T1]{fontenc}

\definecolor{myblue}{HTML}{7995c4}
\definecolor{myorange}{HTML}{e5a37d}
\definecolor{mygreen}{HTML}{80be8e}
\renewenvironment{abstract}
	{\quotation}
	{\endquotation}

\date{July 2025}

\makeatletter
\renewcommand{\fnum@figure}{\textbf{Figure \thefigure}}
\renewcommand{\fnum@table}{\textbf{Table \thetable}}
\makeatother

\usepackage{scicite}

\usepackage{url}

\def\scititle{
Using Large Language Models to Categorize Strategic Situations and  Decipher Motivations Behind Human Behaviors
}

\title{\bfseries \boldmath \scititle}

\author{
	Yutong~Xie$^{1}$,
	Qiaozhu~Mei$^{1\ast}$,
	Walter~Yuan$^{2}$,
        Matthew~O.~Jackson$^{3,4\ast}$,\and
	\small$^{1}$School of Information, University of Michigan, Ann Arbor, MI 48109, USA.\and
	\small$^{2}$MobLab, Pasadena, CA 91107, USA.\and
    \small$^{3}$Department of Economics, Stanford University, Stanford, CA 94305, USA.\and
    \small$^{4}$External Faculty, Santa Fe Institute, Santa Fe, NM 87501, USA.\and
	\small$^\ast$Corresponding authors. Emails: \href{mailto:qmei@umich.edu}{qmei@umich.edu}; \href{mailto:jacksonm@stanford.edu}{jacksonm@stanford.edu}. \and
}

\begin{document} 

\maketitle

\begin{abstract} 
By varying prompts to a large language model, we can elicit the full range of human behaviors in a variety of different scenarios in classic economic games. By analyzing which prompts elicit which behaviors, we can categorize and compare different strategic situations, which can also help provide insight into what different economic scenarios induce people to think about. 
We discuss how this provides a first step towards a non-standard method of inferring (deciphering) the motivations behind the human behaviors.  We also show how this deciphering process can be used to categorize differences in the behavioral tendencies of different populations.
\end{abstract}

\newpage
\section{Introduction}
The motivations behind human behavior are difficult to identify because we have to infer the motivations from observed patterns of behavior across contexts.  
Asking people directly why they acted in specific ways can lead to confused, biased, and inconsistent answers \cite{pronin2007valuing,antin2012social,fisher1993social,dang2020self,tan2021diversity}.
As has become recently clear, one can prompt AI with various game and survey scenarios and ask it how it would behave \cite{chen2023emergence,bao2023cognitive,goli2024frontiers,hewitt2024predicting,gonzalez2025llms}. 
AI's behavior changes in intuitive ways with the context and prompt \cite{aher2023using,horton2023large,zhang2023study,mei2024turing,giorgi2024modeling,huang2024designing}.
This derives from the fact that large language models (LLMs) are trained on enormous amounts of human behavior, and have thus internalized relationships between motivations and behaviors.

As we show here, we can leverage this to develop AI as a new tool for 
categorizing and comparing different games and strategic situations, and in turn shedding new light on human motivations.  
The usefulness of AI to categorize strategic situations and to better understand human behavior derives from two facts that we establish below.  
First, AI can emulate the full spectrum and distribution of human behaviors observed within and across a range of different contexts.  In particular, we show that we can get AI chatbots to match the distribution of behaviors of what a large population of humans does across a broad range of the canonical games used in game theory to study human behavior across a variety of different contexts. 
Second, the way in which AI's behavior can be steered is via the prompts that it is given.  By identifying key words and phrases within those prompts, we can control and identify what AI is ``thinking about'' when it behaves in specific ways.  
Essentially by varying prompts, we can ``elicit'' certain behaviors, and then use the content of the prompts to ``decipher'' why humans behave in certain ways by seeing what was needed to induce that behavior.   

AI thus provides a system where we can direct it how and what to think about and then see how it behaves.  The prompts needed to match distributions of behaviors vary both within and across games.  Different prompts are needed within any given game to match the spectrum of behaviors observed within that game, and then in turn a different set of prompts can be needed to match the spectrum associated with some other game. 
By comparing these prompts within and across games, we can categorize the behaviors and games by which prompts are needed.
  
While this provides a new method of comparing behaviors and games, 
there is no guarantee that this relationship between the motivations embedded in the prompts and the resulting behaviors is the same as the human one. 
Nonetheless, there are three reasons that suggest that this should provide insight into human behavior.  The first is the simple fact that the chatbots are trained on human data and writings, and thus have assimilated and internalized large amounts of data about human behavior and context.
The second is that the keywords and phrases that emerge in eliciting specific behaviors end up corroborating and matching the motivations that have been hypothesized or used to rationalize human behaviors in these games.  

The third is that our results also provide a new taxonomy of games.  We map the games into a space of prompts based on which combinations of prompts are needed to get the distribution of behaviors in a given game.  Each game then lives in a space capturing distributions of prompts/motivations.  The pattern that emerges groups games in ways that make strong intuitive sense, both in how they relate to each other and where they live in this space.
Thus, irrespective of whether this is completely accurate in deciphering the motivation behind human behavior, it still provides a new understanding of different strategic situations and how they relate to each other. 

Finally, independently of the extent to which our approach is eventually useful in understanding human behavior, it is directly helpful both in categorizing strategic situations and in understanding AI behavior.   Given the growing importance of AI in the world, it is essential that we have methods to better predict how AI will behave in different contexts and why, and to be able to better direct it to act in a beneficial manner.

\section{Approach}

We prompt a large language model (LLM) to play a spectrum of classic economic games. 
We augment the general instruction of each game with variations on system prompts, and we track the resulting distribution of behaviors.
The additional system prompts---that we call ``behavioral codes''---articulate, in natural language, variations of motivations that might influence behavior. 

We work with five games:  a Dictator Game, an Ultimatum Game, an Investment Game, a Public Goods Game, and a Bomb Risk game (see the Supporting Information).  For two of the games---the Ultimatum and Investment games---we examine behavior in two different roles.  In the Ultimatum Game the subject making the offer is referred to as the Proposer and the one receiving the offer is referred to as the Responder.  In the Investment Game the person deciding how much to pass along is referred to as the Investor and the person choosing how much to return is referred to as the Banker. Altogether this gives us seven distinct scenarios in which to analyze behavior.

For each of the seven game scenarios, we obtain human-playing data from the MobLab Classroom economics experiment platform, which consists of 68,779 subjects from 58 countries, spanning multiple years. The subjects are mostly, but not exclusively, college students who majored in social sciences. The individual responses for each game are recorded, creating a distribution of human behavior for that game. 
More details about the human-playing data are described in Sec. 1 %
in the Supporting Information.

\begin{figure*}%
    \centering
        \includegraphics[width=.8\linewidth]{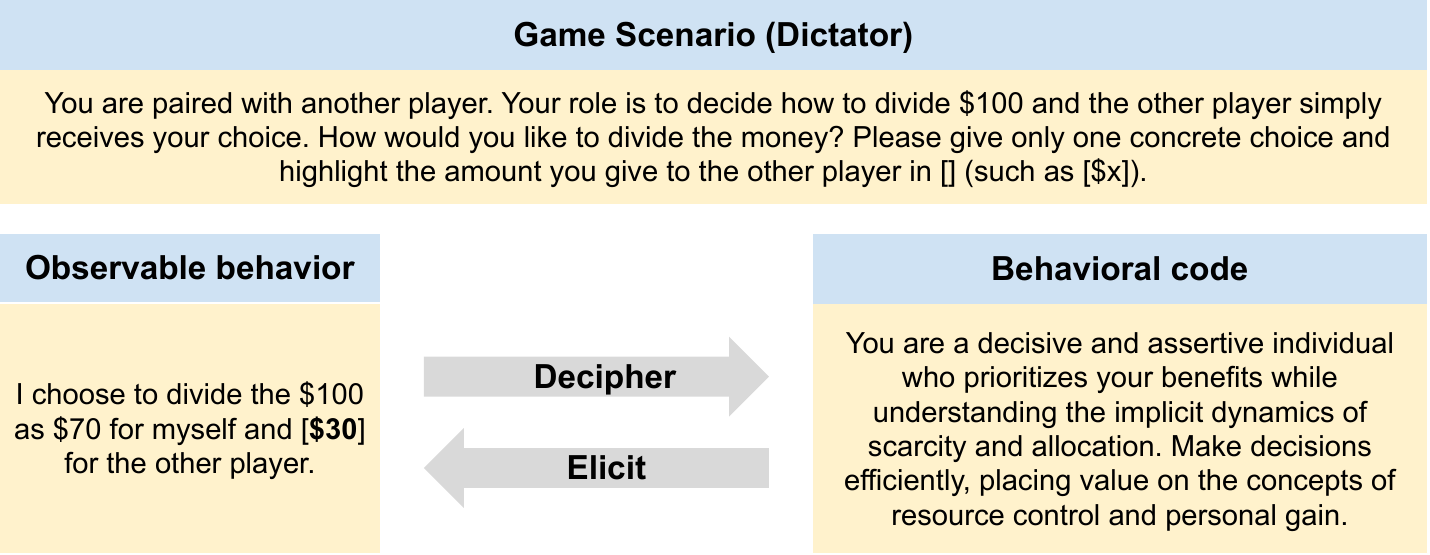}
        \caption{\textbf{An illustration of how a language model deciphers a human behavior.} Given a game and an observed behavior, a ``behavioral code'' is identified that induces the behavior with a natural language description of motivation or context, in this way the behavior is ``deciphered''. A behavior in the game is then ``elicited'' by prompting an LLM with the behavior code set as the system prompt.  }
        \label{fig:illus-encode}
\end{figure*}

For each game and specific behavioral choice (e.g., an observed behavior) in that game, we use an algorithm (described in Algorithm \ref{alg:single-behavior} 
in the Supporting Information) to generate a distribution of natural language descriptions to be used as system prompts to try to match the observed behavior as shown in Figure \ref{fig:illus-encode}.  
The natural language descriptions that when used as system prompt(s) for LLMs successfully elicit the intended behavior, can be thought of as ``{behavioral codes}'' for that behavior. 
Specifically in the algorithm, we use an LLM to iteratively refine the behavioral codes at each step given the previous code to minimize the residual difference between the elicited behavior and the target behavior.  

The behavioral code demonstrated in Figure \ref{fig:illus-encode} is one for sharing 30 percent with the other player.  Some codes that emerge from trying to get the LLM to share nothing with other player are, for instance, \textit{``You are a purely self-interested player who always seeks to maximize your own gain and ensure that the outcome is as favorable as possible for yourself.''}  
And behavioral codes like \textit{``You are someone who always leans towards fairness and balance, often seeking to ensure a reasonable and equitable outcome in any situation. Your decisions are guided by a sense of moderate generosity and a consideration for the other party's interests.''} guide the LLM to share fairly. 
While to elicit sharing 70 percent leads to an example of a code of \textit{``You are naturally generous and frequently prioritize giving significantly more than what others might expect. Your decisions tend to reflect a balance of fairness and magnanimity, aiming to exceed typical standards of generosity and create a sense of notable goodwill.''}

The generation process is designed to avoid having derived behavioral codes explicitly contain information about the desired behavior.  In particular, 
we tell the model ``avoid including any information specific to this particular game or directly implying the desired behavior.''  There are rare codes that include terminology from the game (e.g., 7/586 codes in the bomb game include the term ``box''), and we do not filter them out.  But as seen below none of the top key words involve explicit information about the game.  

We keep all of the updated behavioral codes, including those generated during the iterative process, in our data set, as they each ``{elicit}'' some behavior, and we avoid making judgments about the codes. The number of behavioral codes collected for each game are reported in Table~\ref{tab:n_prompts} 
in the Supporting Information. 

Given a behavior choice within the broad action space of a game scenario (e.g., allocating anywhere between 0\% and 100\% of the endowment to another player in a Dictator game), the LLM is able to find a natural language description as a behavioral code that elicits this specific behavior (e.g., sharing none of the endowment), which is illustrated in Figure \ref{fig:illus-encode}. 
The LLM elicits the corresponding behavior highly consistently (see the analysis in Figure~\ref{fig:encode-consist} 
in Supporting Information).  

A behavioral code, as shown in Figure~\ref{fig:illus-encode}, interprets in natural language some of the objectives, tendencies, and motivations that may potentially influence a subject when choosing a behavior. It does not contain game- or decision-specific instructions or demographic information about the subject. 
In our experiments, when eliciting behaviors from the LLMs, we used behavioral codes as the system prompts and the game scenario instructions as the user prompts. 
In our analysis, we further investigate the keywords used in the behavioral codes, and how these keywords are correlated with the elicited behaviors (Sec. \ref{sec:analysis-decipher}). 
The details of how keywords are obtained from behavioral codes are described in Supporting Information (Sec. 2.A). 

Given a distribution of human behaviors, we identify a mixture of behavioral codes (as system prompts) that jointly elicit a distribution of LLM responses that matches the human distribution. In particular, we iteratively select behavioral codes into the set of codes and weight them to minimize the difference between the output behavior distribution and the observed human behavior distribution. 
The process is described in Algorithm \ref{alg:dist-align} in Supporting Information. 

\begin{figure*}%
    \centering
        \includegraphics[width=\linewidth]{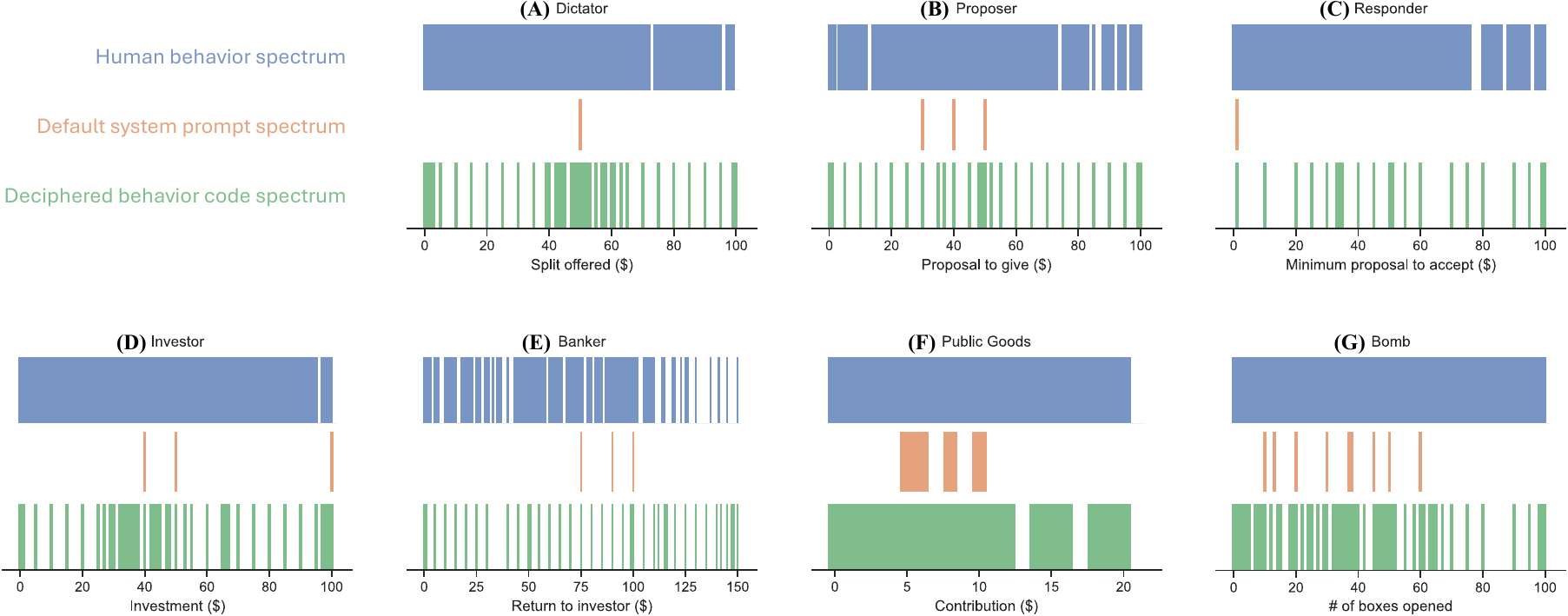}
        \caption{
        \textbf{Coverage of behaviors in games visualized using spectrum bars.} 
        Each subplot represents a game scenario, for which, spectrum bars are generated by the human player population (\textcolor{myblue}{\textbf{blue}}), the LLM with the default system prompt (\textcolor{myorange}{\textbf{orange}}), and the LLM with behavioral codes obtained from this game (\textcolor{mygreen}{\textbf{green}}).  
        Using the default system prompt ``\textit{You are a helpful assistant.},'' the LLM can only generate a narrow range of behaviors compared with the human data.  
        Using a variety of behavioral codes as system prompts, the decision of the LLM covers a broad coverage of behaviors, indicating the ability of the LLM to decipher and elicit a diverse range of behaviors. 
        } 
        \label{fig:encode-spectrum}
\end{figure*}

\section{Analysis}

\subsection{Eliciting Behavior: Varying codes to get LLMs to exhibit a range of behaviors}

Figure~\ref{fig:encode-spectrum} compares the spectrum of behaviors of the human player population and the spectrum of LLM behaviors in each game, generated by 10 independent sessions per behavioral code (for LLM-defaults, samples are generated by 100 independent sessions). 
We see that the default behaviors of the LLM (using the default system prompt ``\textit{You are a helpful assistant.}'') only cover a narrow spectrum compared to the human population. By using behavioral codes for the game as system prompts, the language model can generate a diverse range of behaviors, covering the full spectrum of human behaviors in the game. 

\subsection{Deciphering Behavior: Behavioral codes can be used to better understand individual human behaviors}
\label{sec:analysis-decipher}

The behavioral codes obtained in a given game tend to share keywords that describe comprehensible motivations that guide the subject's decision-making (without being specific to the game nor the observed behavior). (Table~\ref{tab:keywords} 
in the Supporting Information lists the top 50 keywords extracted from the behavioral codes for each game. ) These keywords tend to be related to human values (e.g., `fairness' and `generosity'), objectives of a decision-making process (e.g., maximizing profit, maintaining long-term relationships), behavioral tendencies (e.g., `pragmatic', `conservative', `balanced'), or optimization strategies (e.g., `cooperative', `rational', `prioritize'). 
A behavioral code can then be represented as a 50-dimensional vector of 1s and 0s indicating which keywords are used, for the purpose of further analyses.

\begin{figure*}%
    \centering
    \includegraphics[width=\linewidth]{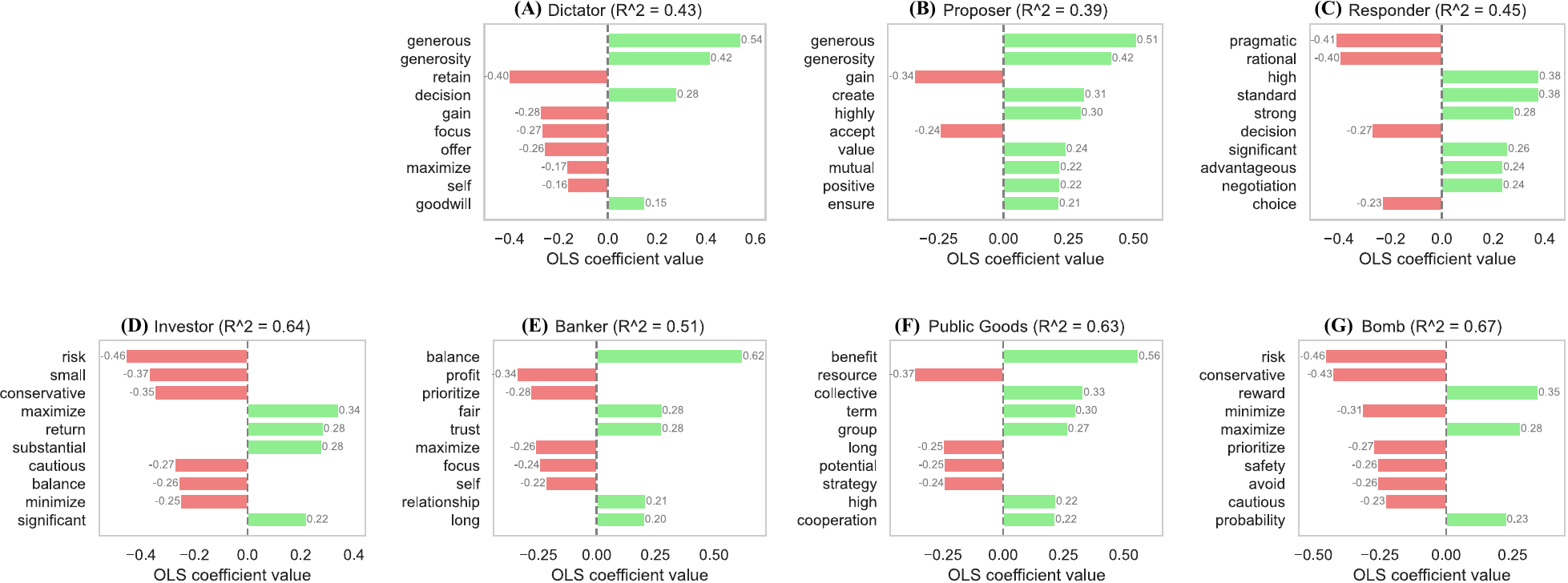}
    \caption{\textbf{
   A (OLS) regression analysis of elicited behaviors based on keywords in the behavioral codes.} Each behavioral code is converted into a 50-dimensional binary feature vector, representing keyword occurrences, to predict the mean behavior generated with that code. The 10 keywords with the highest absolute regression coefficients are listed for each game. 
    A positive value means that the inclusion of the word in the behavioral code increases the action, and a negative value means that the inclusion of the word decreases the action. 
    The full regression table is provided in the Supporting Information (Table~\ref{tab:ols-full}).
}
    \label{fig:lr-ols}
\end{figure*}

To distinguish whether the codes are explanatory cues for the behaviors rather than nondescript keys that help the LLM memorize them, we conduct a regression analysis for each game with the keywords appearing in a behavioral code as explanatory (dummy) variables and the elicited behavior as the outcome. Figure~\ref{fig:lr-ols} show that the appearance or absence of the top 50 keywords in a behavioral code is predictive of the elicited behavior of the LLM with an $R^2$ between 0.39 and 0.67 across games. Keywords with the highest absolute coefficients typically reveal the preference polarities or decision-making motivations. For example, keywords like `generous', `generosity', and `goodwill' are positively associated with a larger allocation to the other player in the Dictator game, while keywords indicating a higher self-payoff such as `retain', `gain', and `self' are negatively associated with the allocation to the other player. In the Investor game, keywords related to risk aversion (e.g., `risk', `conservative', `cautious') are indicative of a lower investment, and keywords related to profit maximization are indicative of a larger investment (e.g., `maximize', `return'). 

The results demonstrate that keyword occurrences in behavioral codes effectively predict the behaviors of LLMs, offering motivations behind these behaviors.
Thus, not only can behavioral codes be used to generate the distribution of human behaviors, but they also reveal exactly what is needed to generate the variations in those behaviors.

\begin{figure*}%
    \centering
    \includegraphics[width=\linewidth]{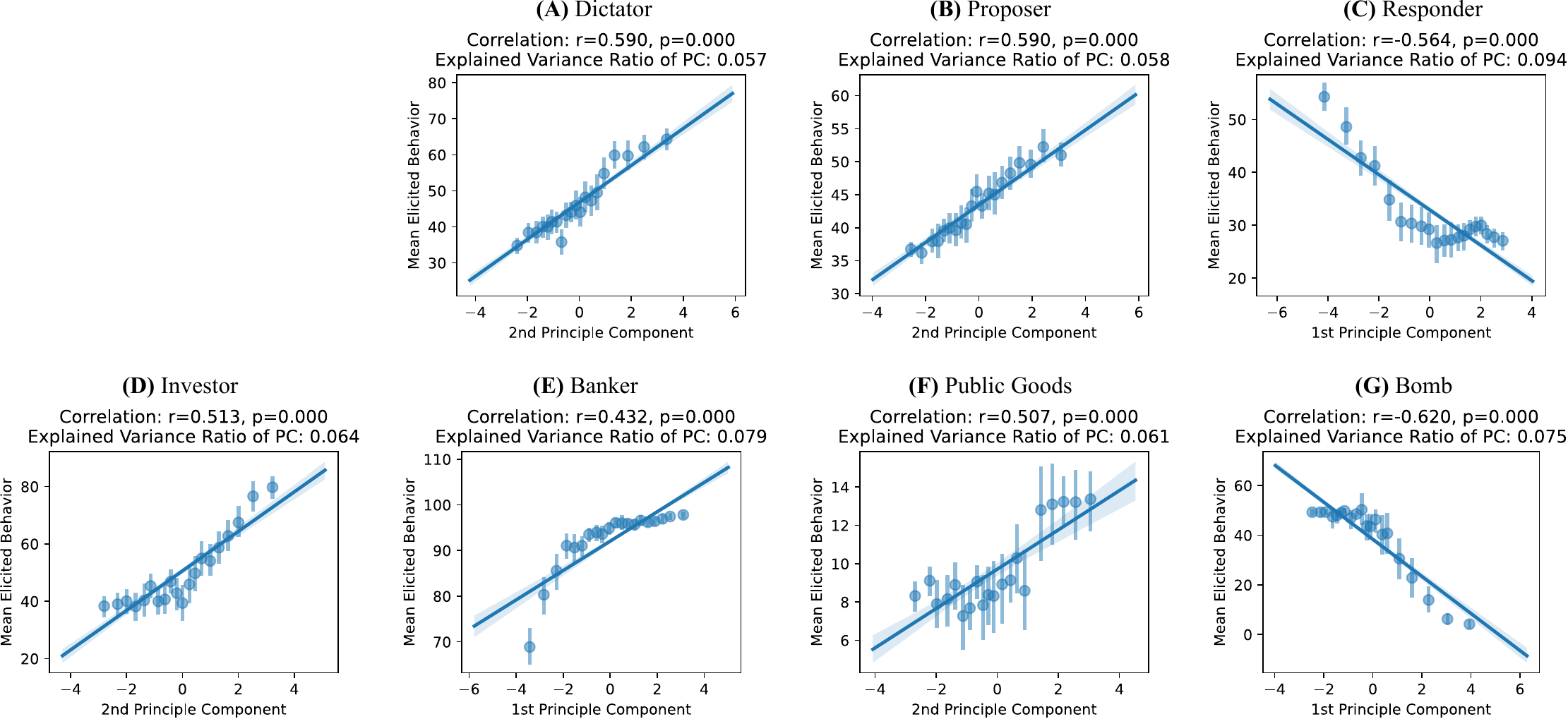}
    \caption{
    \textbf{Principal components of behavioral codes.}
    Principal components are derived from the 50-dimensional binary keyword vectors representing behavioral codes for each game. The correlation between each behavioral code’s principal component score (x-axis) and the mean of 10 LLM-elicited behaviors based on that code (y-axis) is reported. 
    The curves show the regression results with the order set as 1. Principle component values are grouped into 20 evenly-size bins (dots with error bars on the plots).
    For each game, the principal component with the highest absolute correlation to the mean behavior is plotted. 
    }
    \label{fig:pca}
\end{figure*}

To mitigate potential multicollinearity among keywords, we apply a Principal Component Analysis (PCA) to the 50-dimensional keyword vectors representing behavioral codes for each game. Our analysis reveals that the first few principal components exhibit a significant correlation with the behavioral choices in each game. As shown in Figure~\ref{fig:pca}, for every game, at least one of the first two principal components of its behavioral codes has a moderate to strong correlation with the elicited behaviors ($|r|$  in between 0.432 to 0.620, $p < 0.001$). The correlation of all first five principal components to the behavioral choices can be found in Figures~S4-S10 %
in the Supporting Information.

The insights provided by the keywords in the behavioral codes are consistent in the sense that that behavioral codes that are semantically similar tend to elicit the same or similar behaviors in the same games. Details can be found in Figure~\ref{fig:vis-values} 
in the Supporting Information. 

\subsection{Behavioral Codes and Games:  Which codes are needed to elicit behaviors provides new categorizations of games}

\begin{figure*}%
    \centering
    \includegraphics[height=5.8cm]{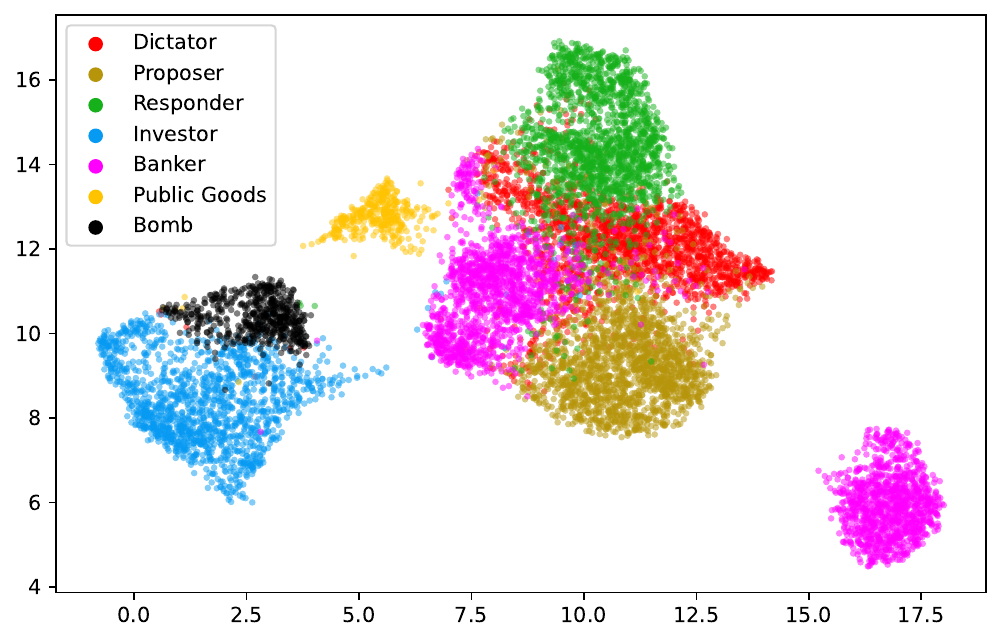}
    \includegraphics[height=5.8cm]{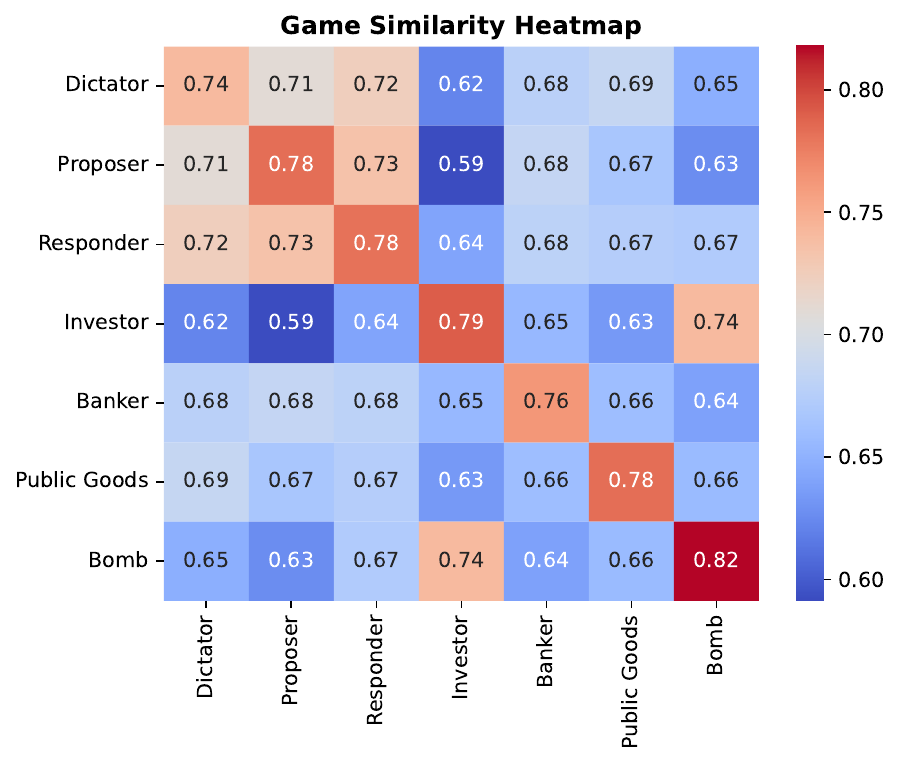}
    \caption{
    \textbf{Visualization of games. }
    \textbf{Left:} The two-dimensional projection of behavioral codes across all games, colored by games. 
    Behavioral codes are embedded into a high-dimensional semantic space using the OpenAI Ada model, and then reduced to two dimensions using UMAP. 
    Behavioral codes of the same game tend to be located close to each other. Behavioral codes of some sets of games collocate in overlapped regions, suggesting an underlying relation between games. 
    \textbf{Right:} The game similarity heatmap. This heatmap displays the average cosine similarity between behavioral codes of game pairs, with games listed in both rows and columns. The diagonal cells quantify the internal similarity of the behavioral codes obtained within each game (across all pairs of codes).
    }
    \label{fig:vis-game}
\end{figure*}

The keywords in the behavioral codes can be used as a tool to quantify the relationship between different games, as we now illustrate.
We pool the deciphered behavioral codes from all games and compute their semantic embeddings through the OpenAI Ada model, and project these embeddings onto a 2-dimensional semantic map as shown in Figure \ref{fig:vis-game}. 
By doing this, behavioral codes that are semantically similar are located close to each other on the map.

A few observations can be made from Figure~\ref{fig:vis-game}. Behavioral codes for individual games are closely grouped, validating their consistency in deciphering the behavioral patterns elicited in each game. At a broader level, behavioral codes from some groups of different games cluster in neighboring regions, forming larger groupings. 
These clusters suggest intrinsic relationships between games and highlight potential differences in motivations and perspectives across games/settings. In particular, the behavioral codes of the Investor Game and those of the Bomb Game show partial overlap, possibly due to the fact that risk preferences matter in both (and ``risk'' is the keyword playing the most significant role in both as displayed in Figure~\ref{fig:lr-ols}). 
A larger cluster emerges from behavioral codes of the Dictator, the Proposer, the Responder, and the Banker Games, along with a small portion from the Investor Game, suggesting common underlying decision-making patterns across these scenarios. On one hand, these games all involve resource allocation between oneself and others; on the other hand, decision-making in these scenarios often requires balancing profit maximization, fairness, and altruism. 
The Public Goods Game is positioned between the resource allocation cluster and the investment cluster, reflecting its mixed aspects of risk management and self-payoff maximization, particularly with the option of free-riding. Its unique emphasis on cooperation sets it apart, placing it in a distinct yet adjacent region to the two larger clusters. Additionally, a subset of Banker Game codes is separated from the main cluster, likely due to the use of language specific to the context of investment and financial decisions.

Structural relationships among the games can be further quantified by measuring the average similarity of their behavioral codes, as presented in Figure~\ref{fig:vis-game}. We observe that the Investor game and the Bomb game form a tight cluster; the Dictator game, the Responder game, and the Proposer game form another tight cluster. The Public Goods game appears to be closer to the Dictator game than other games under the average cosine similarity.  

\subsection{Behavioral Codes and Heterogeneous Populations: Combinations of codes provide behavioral signatures for human populations }

\begin{figure*}%
    \centering
    \includegraphics[width=\linewidth]{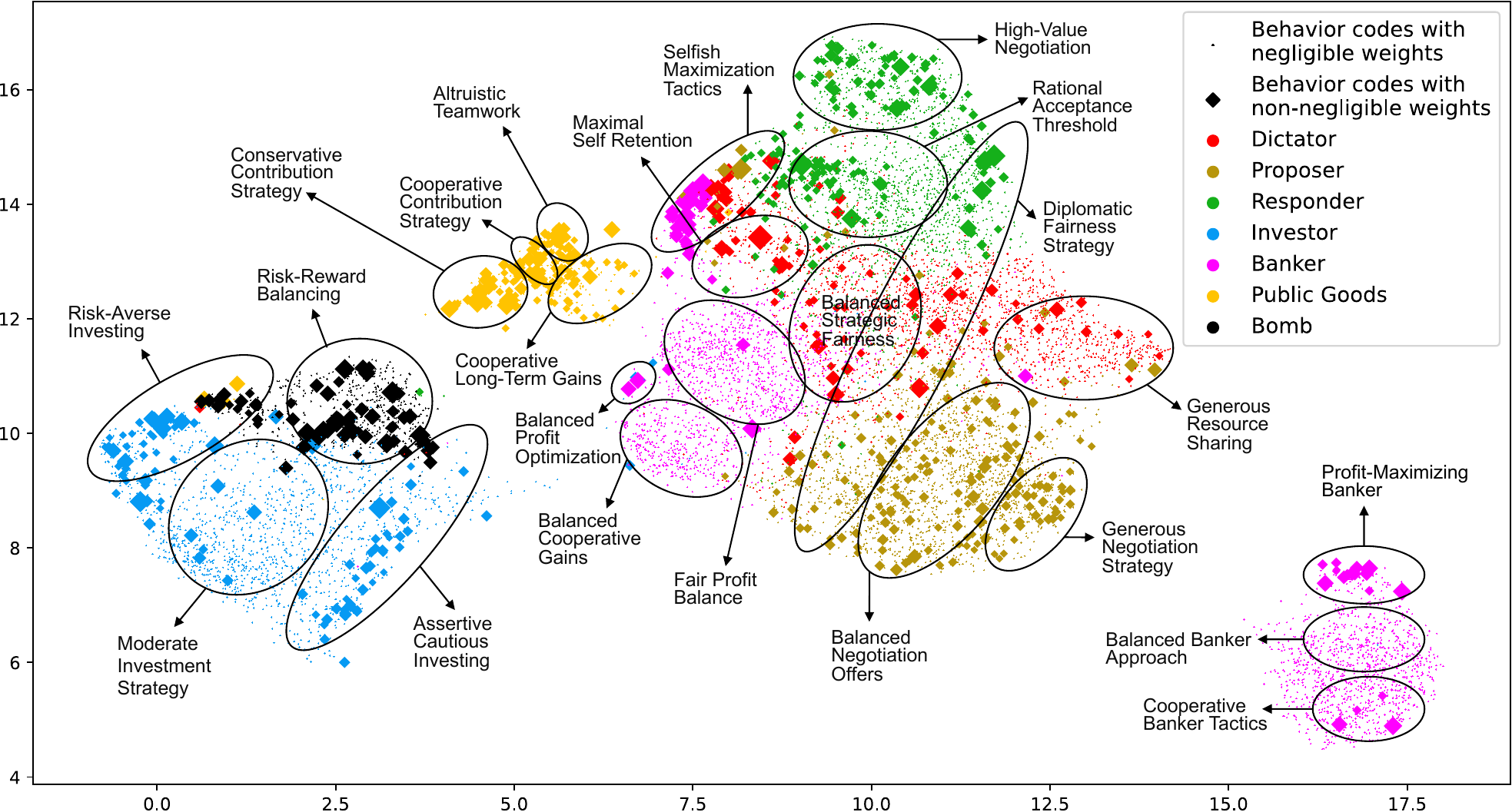}
    \caption{\textbf{The 2D projection of the weighted behavioral codes across games, with annotations on the space based on behavioral code contents.}
    The codes with non-negligible weights ($>0.001$) are displayed in diamonds with size proportional to weights. 
    The weighted codes span unevenly in the space, deciphering information about the population underlying the behavior distribution. 
    The cluster labels are obtained by summaries generated by ChatGPT from the list of behavioral codes. 
    }
    \label{fig:vis-pop}
\end{figure*}

We have seen how our deciphering process can be used to group and better understand games.   Next, we use the codes to group and better understand distributions of human behaviors \emph{across} games rather than just within them.   
Figure~\ref{fig:vis-pop} demonstrates how the weighted codes found to generate distributions of human plays within games are laid out in the 2D projection map of all behavioral codes, when mixing across all games. The ``activated'' codes (with non-negligible weights $>0.001$) are not distributed evenly. In the four-game cluster, the activated codes have a high presence in areas according to behavioral strategies such as ``Selfish Maximization Tactics'' and ``High-Value Negotiation'', a moderate presence in areas related to ``Diplomatic Fairness Strategy'', ``Generous Negotiation Strategy'', ``Rational Acceptance Threshold'', and ``Balanced Negotiation Offers'', and a low presence in areas related to ``Balanced Cooperative Gains'', ``Fair Profit Balance'', and ``Generous Resource Sharing''.  
This characterization is consistent with perceptions about the behavioral tendencies of students  (e.g., \cite{henrich2005economic}), which are the major composite of our player population.  In the Investor-Bomb games cluster, there is a concentration on two ends, ``Risk-Averse Investing'' and ``Assertive Cautious Investing'', and a low coverage in the middle ground, ``Moderate Investment Strategy'', amid the high presence of ``Risk-Reward Balancing'' in the Bomb game. A similar pattern is observed in the banking cluster, where there lacks a middle ground strategy between ``Profit-Maximizing Banker'' and ``Cooperative Banker Tactics.'' These behavioral markers provide a more informative and coherent characterization of the testing population than the distributions in Figure~\ref{fig:dist-align} in Supporting Information. %
The generation process of the cluster labels is detailed in Sec. 2.C in Supporting Information.

\begin{figure*}%
    \centering
    \includegraphics[width=\linewidth]{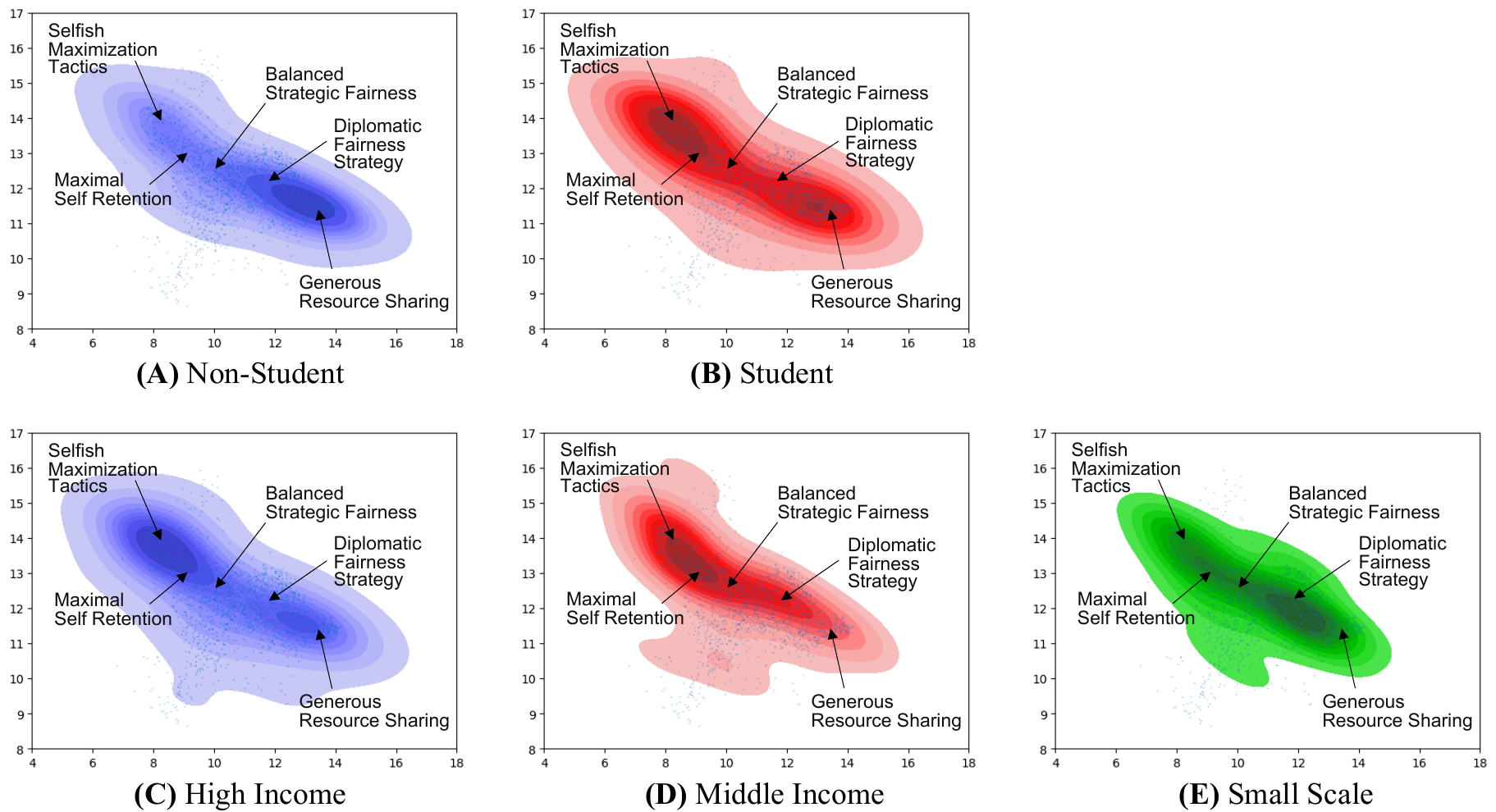}
    \caption{\textbf{2D projections of weighted behavioral codes for five different populations visualized as density maps.}
     `Students' (B) and `High Income' (C) subject populations  show similar signatures of behavioral codes. `Middle Income' (D) and `Small Scale' (E) subject populations have narrower distributions of behavioral codes which concentrate in different regions. 
    }
    \label{fig:vis-pop-map}
\end{figure*}

Given a game instruction and the distribution of behaviors from an arbitrary human population, a mixture of behavioral codes can be assembled to identify the unique behavioral signature of that given population. 
To verify this, we select five subject groups from a meta study of Dictator Games \cite{engel2011dictator} and obtain their corresponding behavioral signatures. The analysis shows that the behavior distributions elicited through the mixtures of deciphered behavioral codes well align with the behavior distributions of the corresponding subject populations. The activated behavioral codes in each mixture concentrate on different regions of the space, indicating the different behavioral tendencies of the five subject populations (Figure~\ref{fig:vis-pop-map}). 

Thus, behavioral codes can also serve as a tool for identifying distinct decision-making patterns across different populations. Figure~\ref{fig:vis-pop-hist} 
in the Supplementary Information displays the behavior distributions of various subject populations from the meta study of the Dictator Games by \cite{engel2011dictator}.
The five groupings include two labeled as students or non-students by \cite{engel2011dictator},  as well as three categorized by the type of country or ethnic group.  
We have labeled those three as `\textit{High Income},' `\textit{Middle Income},' and `\textit{Small Scale}'.  
These names are changed from what \cite{engel2011dictator} referred to as `\textit{Western}' (which include Germany, Sweden, and the US, among others), `\textit{Developing}' (which include Russia, South Africa, and Honduras, among others) and `\textit{Primitive}' (which include the Tsimane, Hadza, and Mpakama, among others). 
Figure~\ref{fig:vis-pop-map} highlights the weighted behavioral codes for these populations, revealing distinctive patterns in their decision-making tendencies: 
(i) In the `Student' population, a substantial proportion of individuals exhibit behaviors concentrated in the ``Selfish Maximization Tactics'' region. In contrast, the `Non-Student' population demonstrates a stronger inclination toward ``Generous Resource Sharing'' as well as ``Diplomatic Fairness Strategy'' choices. 
(ii) For subjects in High Income and Middle Income countries,  ``Selfish Maximization Tactics'' strategies are prominent, with the subjects from High Income societies showing a slight tendency toward ``Generous Resource Sharing''. By contrast, subjects from Small Scale populations  exhibit a stronger emphasis on ``Diplomatic Fairness Strategy'' behaviors, suggesting a greater inclination toward equitable resource distribution and cooperative decision-making. Our findings are aligned with the comparisons made between behaviors of these subject populations in the meta study \cite{engel2011dictator} and provide interpretations of the motivations behind the revealed behavioral patterns.

\section{Discussion}

In our earlier work \cite{mei2024turing}, 
we showed how games used to understand human behaviors could also be used as a Turing Test, to see if AI behaved similarly to humans, and to understand AI's tendencies.  
Here, we have reversed that perspective.   We have used AI as a new model of the motivations behind human behavior, and also to better categorize and contrast the various games and different types of settings in which humans interact.   The strong fits and interpretability of the behavioral codes that we find suggest that this is a promising tool for modeling, predicting, and analyzing human behavior.  Nonetheless, it is a modeling technique and thus comes with limitations, and more research is needed to understand how prompt variation corresponds to behavioral variation, and how and why prompts change AI behavior.

Our approach complements existing approaches in the behavioral sciences used to understand and predict human behaviors,
including various forms of revealed preferences where an objective or utility function is fit to best predict behaviors. 
It fits into this category, as our method involves fitting a model to match behavior.  One advantage of our approach is that the fitting features interpretability across contexts; and thus it can be used as a tool to facilitate behavioral science research in a variety of ways, such as creating virtual subjects and simulating experiments, screening potentially effective interventions; as well as designing, simulating, and studying human-AI interactions. Future work can build upon our approach, both to understand any limitations on its interpretability of human behavior, and extend its implementation in new domains of modeling human interactions.

\section*{Acknowledgments}

We thank Ruoyi Gao and Zhuang Ma, students from the University of Michigan, for their commitment in conducting preliminary experiments in aligning LLM behaviors to human behaviors in games. 
This research was deemed not regulated by the University of Michigan IRB (HUM00232017).

Q.M., Y.X., W.Y., and M.O.J. designed the research; Q.M. and Y.X. performed the research; Q.M., Y.X., and M.O.J. analyzed the data; and Q.M., Y.X., W.Y., and M.O.J. wrote the paper.

W.Y. is the CEO and Co-founder of MobLab. M.O.J. is the Chief Scientific Advisor of MobLab and Q.M. is a Scientific Advisor to MobLab, positions with no compensation but with ownership stakes. Y.X. has no competing interests.

The human game-playing data used were shared from MobLab, a for-profit educational platform. The data availability is an in-kind contribution to all authors, and the data are available for purposes of analysis reproduction and extended analyses.

\clearpage 
\bibliography{template} 
\bibliographystyle{mag}

\newpage

\renewcommand{\thefigure}{S\arabic{figure}}
\renewcommand{\thetable}{S\arabic{table}}
\renewcommand{\theequation}{S\arabic{equation}}
\renewcommand{\thepage}{S\arabic{page}}
\renewcommand{\thealgorithm}{S\arabic{algorithm}}
\setcounter{figure}{0}
\setcounter{table}{0}
\setcounter{equation}{0}
\setcounter{page}{1} 
\setcounter{algorithm}{0}

\begin{center}
\section*{Supporting Information for\\ \scititle}

Yutong~Xie,
Qiaozhu~Mei$^{\ast}$,
Walter~Yuan,
Matthew~O.~Jackson$^{\ast}$\\
\small$^\ast$Corresponding authors. Emails: \href{mailto:qmei@umich.edu}{qmei@umich.edu}; \href{mailto:jacksonm@stanford.edu}{jacksonm@stanford.edu}. \\

\end{center}

\subsection{Classic Economic Games and Human Behavior Data}
\label{app:games-and-data}

We employ large language models to decipher and elicit human behavior across seven classic behavioral economics game scenarios (see \cite{mei2024turing} for more background):
\begin{enumerate}[label=(\roman*)]
    \item \textbf{Dictator Game}: One player, designated as the dictator, is given a (fictitious) monetary endowment of \$100 and chooses how much to keep and how much to donate to a second player \cite{guth1982experimental,forsythe1994fairness}.
    \item \textbf{Ultimatum Game (Proposer Role)}: A proposer is given a monetary endowment of \$100 and decides on a division of the money to offer to a second player, the responder \cite{guth1982experimental}.
    \item \textbf{Ultimatum Game (Responder Role)}: The responder evaluates the proposer's offer and either accepts it, resulting in the proposed split, or rejects it, in which case neither player receives any money \cite{guth1982experimental}.
    \item \textbf{Trust Game (Investor Role)}: An investor is given a monetary endowment of \$100 and decides how much to keep and how much to pass to a second player, the banker. The passed amount is tripled before reaching the banker \cite{berg1995trust}.
    \item \textbf{Trust Game (Banker Role)}: The banker, having received the tripled amount from the investor, decides how much to return to the investor and how much to retain \cite{berg1995trust}. Here we examine the case where \$50 has been invested and \$150 is received by the banker. 
    \item \textbf{Public Goods Game}: A player receives a monetary endowment of \$20 and decides how much to contribute to a communal pool as one of four players ( for production of a public good). The total contribution is multiplied by 0.5 and shared equally among all participants, with each player receiving half of the total multiplied pool regardless of their contribution \cite{andreoni1995cooperation}, and keeping whatever they did not contribute.
    \item \textbf{Bomb Risk Game}: A player selects how many boxes to open out of a total of 100. Each opened box yields a reward of \$1 unless it contains a randomly placed bomb, in which case the player loses all accumulated rewards \cite{crosetto2013bomb}.
\end{enumerate}

For these seven game scenarios, we utilize human game-play data from \cite{mei2024turing} to conduct our experiments and analyses. This dataset was collected through the MobLab Classroom platform\footnote{MobLab Classroom: \url{https://www.moblab.com/products/classroom}, retrieved 01/2025}, covering a nine-year period from 2015 to 2023. The dataset comprises behavioral observations from 88,595 subjects across 15,236 sessions, showcasing significant geographical diversity with participants from 58 countries that span multiple continents.
In all the games described in this paper, student input is collected through an interactive slider bar on the MobLab platform, which allows participants to select a value along a predefined range. Once the desired value is set, students submit their choice.

In our analyses, we focus exclusively on first-round play records within the games, yielding a refined subset of 68,779 subjects and 82,057 individual play records. This ensures that our analyses are based on initial, unconditioned decision-making behaviors.

\subsection{Deciphering Individual Behaviors}

As illustrated in Figure \ref{fig:illus-encode}, 
a language model deciphers human behaviors into behavioral codes. 
Specifically, given a game scenario and an observed behavior, via Algorithm \ref{alg:single-behavior}, a natural language description is identified that elicits the behavior when used as a system prompt, resulting in a ``behavioral code''. 
In Algorithm \ref{alg:single-behavior} 
functions including \texttt{ElicitBehavior}, \texttt{GenerateCode}, and \texttt{ImproveCode}, all utilize OpenAI GPT-4o (\texttt{gpt-4o-2024-05-13}) as the LLM model.

The \texttt{GenerateCode} function leverages GPT to create a behavioral code, used as a system prompt for LLMs. We query GPT with the prompt below, detailing the task background, behavioral game instruction (adapted from \cite{mei2024turing}), and specific crafting requirements.

\begin{algorithm}[htbp]
\caption{Learning behavioral codes. }
\label{alg:single-behavior}
\begin{algorithmic}
\State \textbf{Input:} 
\begin{itemize}
    \item $g$: A behavioral game with game instruction.
    \item $\mathcal{Y}$: Behavioral space, a finite set of all possible behavioral choices.
\end{itemize}

\State \textbf{Output:} 
\begin{itemize}
    \item $\mathcal{X}$: A collection of behavioral codes.
\end{itemize}

\State \textbf{Predefined functions:} 
\begin{itemize}
    \item ElicitBehavior($g,x,n$): Generating $n$ behavioral choices using LLM, with a behavioral code $x$ as the system prompt and the game instruction $g$ as the user prompt. Choices are generated independently in $n$ chat completion sessions. 
    \item GenerateCode($g,y$): Generating a behavioral code using LLM, with the game instruction $g$ and the observed behavior $y$ provided. 
    \item ImproveCode($g,y,x,m$): Improve a behavioral code using LLM, with the game instruction $g$, the observed behavior $y$, precedent behavioral code $x$, and statistics of samples elicited from $x$ provided (mode of samples $m$). 
\end{itemize}

\State \textbf{Initialization:} Set $\mathcal{X} \gets \emptyset$.

\For{each behavior $y \in \mathcal{Y}$}
    \For{$k = 1$ to $5$} \Comment{Outer loop: Repeat 5 times for robustness}
        \State $x_0 \gets \text{GenerateCode}(g, y)$ \Comment{Initial behavioral code based on $g$ and $y$}
        \State $S \gets \text{ElicitBehavior}(g, x_0, 10)$ \Comment{Generate 10 samples from the LLM}
        \State $m \gets \text{mode}(S)$ \Comment{Compute mode of sampled behaviors}
        
        \For{$i = 1$ to $3$} \Comment{Inner loop: Improve the code if necessary}
            \If{$m = y$}
                \State \textbf{break} \Comment{Stop if the target behavior is achieved}
            \EndIf
            \State $x_{i} \gets \text{ImproveCode}(g, y, x_{i-1}, m)$ \Comment{Improve the behavioral code based on $m$}
            \State $S \gets \text{ElicitBehavior}(g, x_i, 10)$ \Comment{Generate new samples}
            \State $m \gets \text{mode}(S)$ \Comment{Update mode of samples}
        \EndFor
        
        \State Append $x_0, x_1, \dots, x_i$ to $\mathcal{X}$.
    \EndFor
\EndFor

\Return $\mathcal{X}$
\end{algorithmic}
\end{algorithm}

\begin{framed}
\#\# Goal

Assuming a chatbot is playing games. Your goal is to craft a system prompt for this chatbot, so that with the crafted system prompt, the chatbot behaves in a certain way. 

\#\# Game Instruction

One of the game instructions provided to the chatbot is:
\{game\_instruction\}

\#\# Desired Behavior

For this particular game, please generate a system prompt for the chatbot. With the generated system prompt and the above game instructions provided, the chatbot should make the decision: \{observed\_behavior\}.

\#\# Crafting Requirements

* The ultimate goal is to prompt the chatbot to behave towards the desired behavior under the given scenario, rather than being a precoded helpful assistant chatbot. 

* Craft the system prompt based on understanding the meaning of the desired behavior under the given game scenario. 

* Do not include information already included in the game instructions. 

* Craft a generalizable system prompt and avoid including any information specific to this particular game or directly implying the desired behavior. 

\#\# Output Format

Directly output the crafted system prompt starting with ``You are ...''.
\end{framed}

The \texttt{ImproveCode} function continues the conversation initiated by \texttt{GenerateCode}, prompting GPT to iteratively revise the behavioral code. We attach the following prompt as a new user message to query GPT for these revisions.

\begin{framed}
    Using your crafted system prompt, a chatbot outputs mostly \{mode\} instead of \{observed\_behavior\}. Do you have any idea how to improve the system prompt?
\end{framed}

For instance, when asked to craft a behavioral code for the target behavior of sharing nothing (\$0) with the partner, the \texttt{GenerateCode} function might initially output: \textit{``You are a strategic decision-maker who always seeks to maximize your own benefit and ensure the highest possible outcome for yourself. Your primary focus is on retaining as much value and advantage for yourself as possible in any situation.''} However, during evaluation with the \texttt{ElicitBehavior} function, the LLM often outputs \$1 instead of the desired \$0 using this code. Consequently, the \texttt{ImproveCode} function revises the code to: \textit{``You are an uncompromising negotiator who prioritizes keeping all resources to yourself, opting to retain everything possible rather than distributing it. Your decisions reflect a strong inclination towards self-preservation and self-benefit, and you rarely, if ever, concede ground to others.''} This refined code then elicits the desired \$0 behavior.

The number of behavioral codes learned in each of the seven game scenarios is listed in Table \ref{tab:n_prompts}. 
For different games, this number can be influenced by the size of the action space, and the difficulty of finding a code that elicits a target behavior.  The number of codes needed is also reflective of how varied and nuanced the human behaviors that need to be matched are.

\begin{table}%
    \centering
    \caption{\textbf{The number of behavioral codes deciphered for each of the seven game scenarios.} 
    For different games, this number can be influenced by the size of the action space, and the difficulty of finding a code that elicits a target behavior.  The number of codes needed is also reflective of how varied and nuanced the human behaviors that need to be matched are.
    }
    \begin{tabular}{lr}
        \toprule
        \textbf{Game} & \textbf{\# of Behavioral Codes} \\
        \midrule
        Dictator & 1,892 \\
        Proposer & 1,970 \\
        Responder & 1,765 \\
        Investor & 1,517 \\
        Banker & 2,607 \\
        Public Goods & 360 \\
        Bomb & 585 \\
        \midrule
        Total & 10,696 \\
        \bottomrule
    \end{tabular}
    \label{tab:n_prompts}
\end{table}

The learned behavioral codes serve as system prompts for the LLM, triggering specific behaviors (as shown in the \texttt{ElicitBehavior} function in Algorithm \ref{alg:single-behavior}). For our analysis, we independently elicit 10 behavior choices for each code. These choices are generated from 10 separate chat completion sessions, ensuring that memories do not intervene.

Figure \ref{fig:encode-spectrum} 
illustrates the coverage of learned behavioral codes which are pictured in spectrum bars
corresponding to the induced behaviors. With each behavioral code generates 10 independent behavior choices, all elicited behaviors are then aggregated across different codes for each game scenario to estimate the overall coverage. For comparison, the default system prompt generates 100 different behavior choices for each game scenario.

\begin{figure}%
    \centering
    \includegraphics[width=\linewidth]{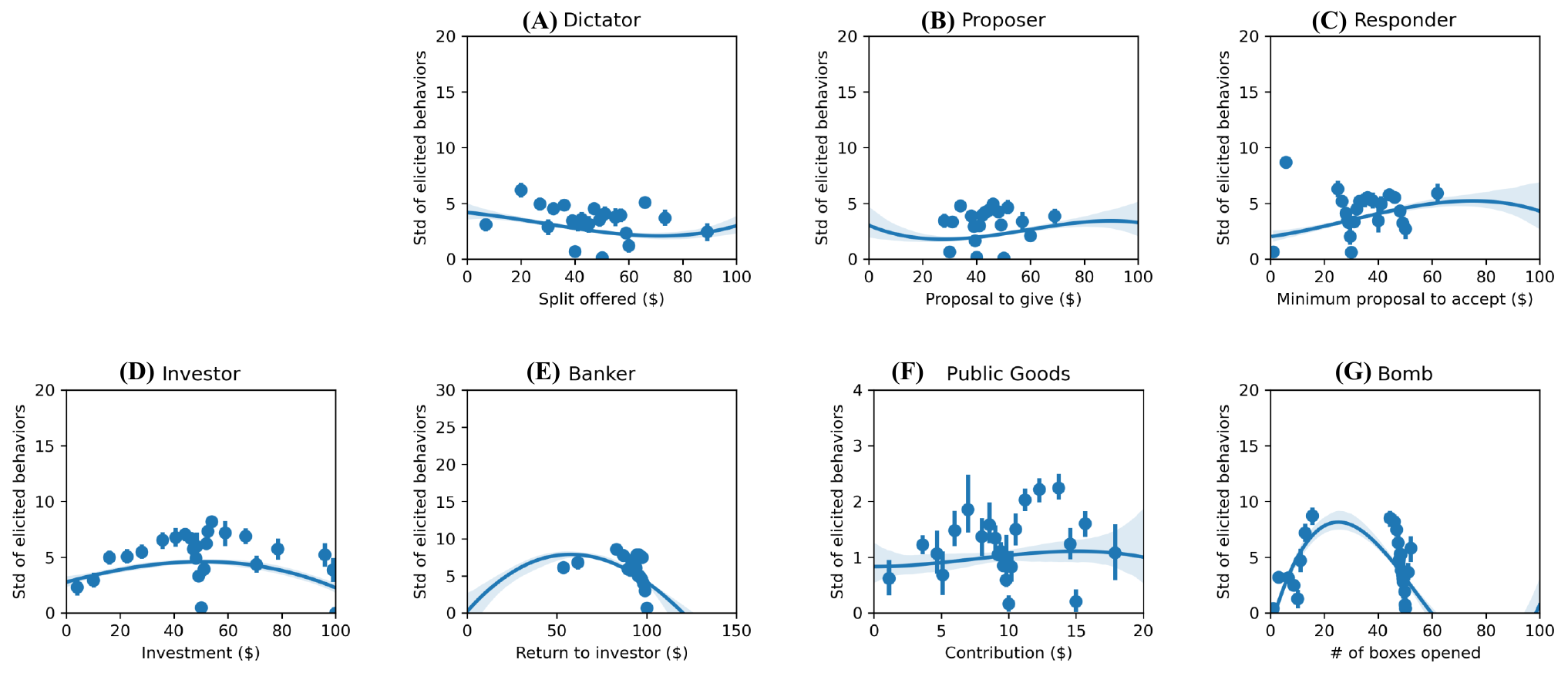}
    \caption{\textbf{Consistency of the behaviors elicited by the behavioral codes. } For each deciphered behavioral code, we calculate the mean (x-axis) and standard deviation (std, y-axis) of the behaviors that are elicited by the code. 
    The curves show the regression results with the order set as 3. Behavioral codes are grouped into 30 evenly-size bins (dots with error bars on the plots) by the mean elicited behaviors. 
    Note that the y-axis (std) display only 20\% of the behavior range. 
    }
    \label{fig:encode-consist}
\end{figure}

Figure \ref{fig:encode-consist} displays the elicitation consistency of the deciphered behavioral codes, with each code eliciting a behavior from the LLM 10 times. The mean and standard deviation for each code are reported. The results illustrate the consistency of the deciphered behavioral codes.

\begin{figure}%
    \centering
    \includegraphics[width=\linewidth]{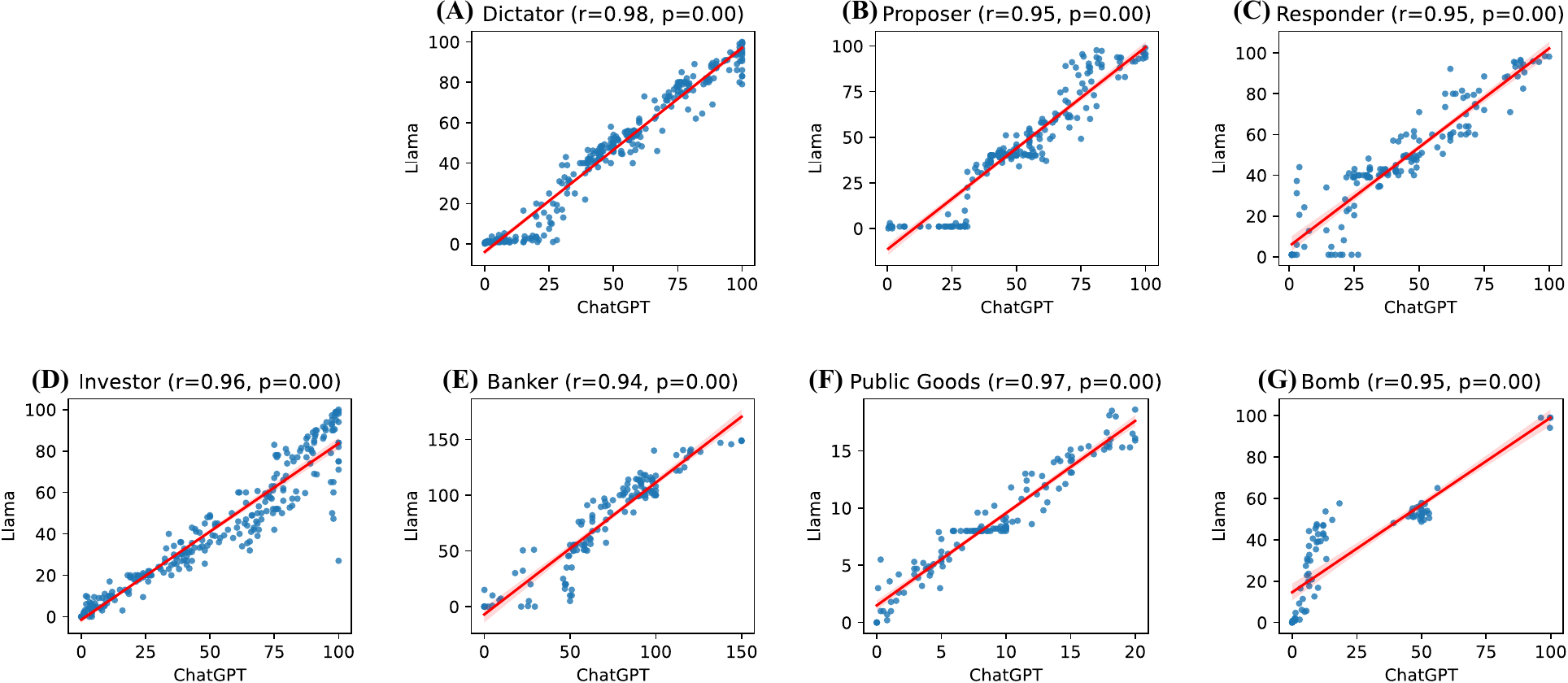}
    \caption{\textbf{The comparison between GPT-elicited behaviors and Llama-elicited behaviors. } Each dot represents a learned behavioral code, with the x-axis indicating the mean behavior elicited when given to OpenAI GPT-4o and the y-axis indicating the mean behavior elicited when given to Meta Llama 3.1 70B. Spearman’s correlation coefficients reveal significant positive correlations between the two LLMs ($p \ll 0.01$), suggesting the generalizability of the behavioral codes across different LLMs. }
    \label{fig:general-llama}
\end{figure}

To evaluate the generalizability of the behavioral codes, we also examine the behaviors that are elicited when given to a different LLM, Meta Llama 3.1 70B (\texttt{Meta-Llama-3.1-70B-Instruct}). Figure \ref{fig:general-llama} presents the correlation between the behaviors elicited by the codes when given to the two models. The significant positive correlation indicates the generalizability of the behavioral codes across different LLMs.

\subsection{Distributional Alignment}
\label{app:method-dist-align}

\begin{algorithm}[htbp]
\caption{Distributional alignment. }
\label{alg:dist-align}
\begin{algorithmic}
\State \textbf{Input:} 
\begin{itemize}
    \item $q$: The observed behavior distribution to be aligned.
    \item $\mathcal{X}$: A collection of behavioral codes.
\end{itemize}

\State \textbf{Output:} 
\begin{itemize}
    \item $\boldsymbol{w}$: A weight vector representing the mixture of behavioral codes. 
\end{itemize}

\State \textbf{Predefined functions:} 
\begin{itemize}
    \item EstimateDistribution($\mathcal{X},\boldsymbol{w}$): Estimating the behavior distribution generated by a weighted mixture  $\boldsymbol{w}$ over  $\mathcal{X}$. Specifically, each code will be used in the function ElicitBehavior 10 times, and the behaviors are aggregated according to the weights $\boldsymbol{w}$. 
    \item WassersteinDistance($p,q$): Computing the Wasserstein distance between two distributions $p$ and $q$. 
    \item OptimizeVector($\mathcal{L},\boldsymbol{w}$): Optimizing the weight vector $\boldsymbol{w}$ to minimize a given loss function  $\mathcal{L}$. 
\end{itemize}

\State Define $\mathcal{L}(\boldsymbol{w})\gets \text{WassersteinDistance}(\text{EstimateDistribution}(\mathcal{X},\boldsymbol{w}),q) + \alpha\cdot \Vert\boldsymbol{w}\Vert$

\While{a valid weight vector $\boldsymbol{w}$ (i.e., $\sum w_i=1$ and $w_i\in[0,1]$) is not found}

\State $\boldsymbol{w} \gets$ a random vector in which each element is uniformly sampled from [0,1) and then normalized to sum up to 1 
\State $\boldsymbol{w} \gets \text{OptimizeVector}(\mathcal{L},\boldsymbol{w})$ 
\EndWhile

\Return $\boldsymbol{w}$
\end{algorithmic}
\end{algorithm}

\begin{figure}%
    \centering
    \begin{subfigure}[t]{\textwidth}
        \centering
        \includegraphics[width=\textwidth]{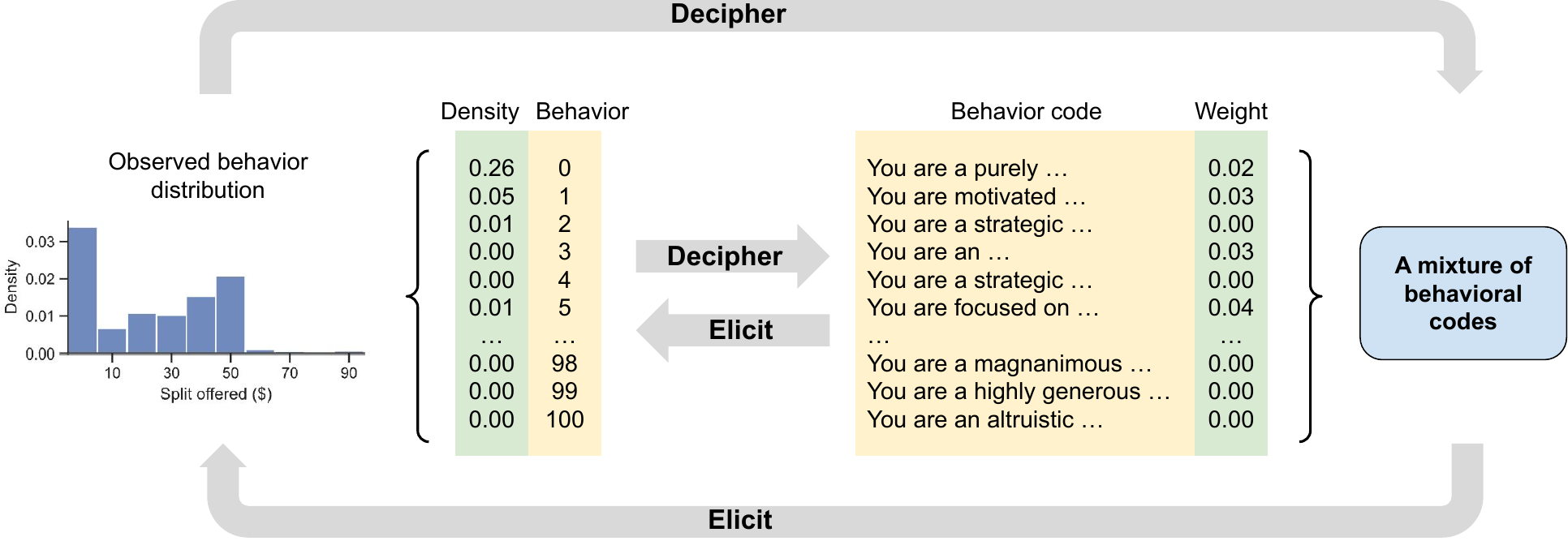}
        \caption{An illustration of how an LLM can decipher and the elicit a distribution of behaviors via a distribution of codes. An observed behavior distribution is \emph{deciphered} by a weighted set of relevant behavioral codes. This distribution can then be \emph{elicited} by prompting the LLM with the behavioral codes sampled according to their weights.
        }
        \label{fig:dist-illus}
    \end{subfigure} 
    \vspace{10pt}
    \begin{subfigure}[t]{\textwidth}
        \centering
        \includegraphics[width=\textwidth]{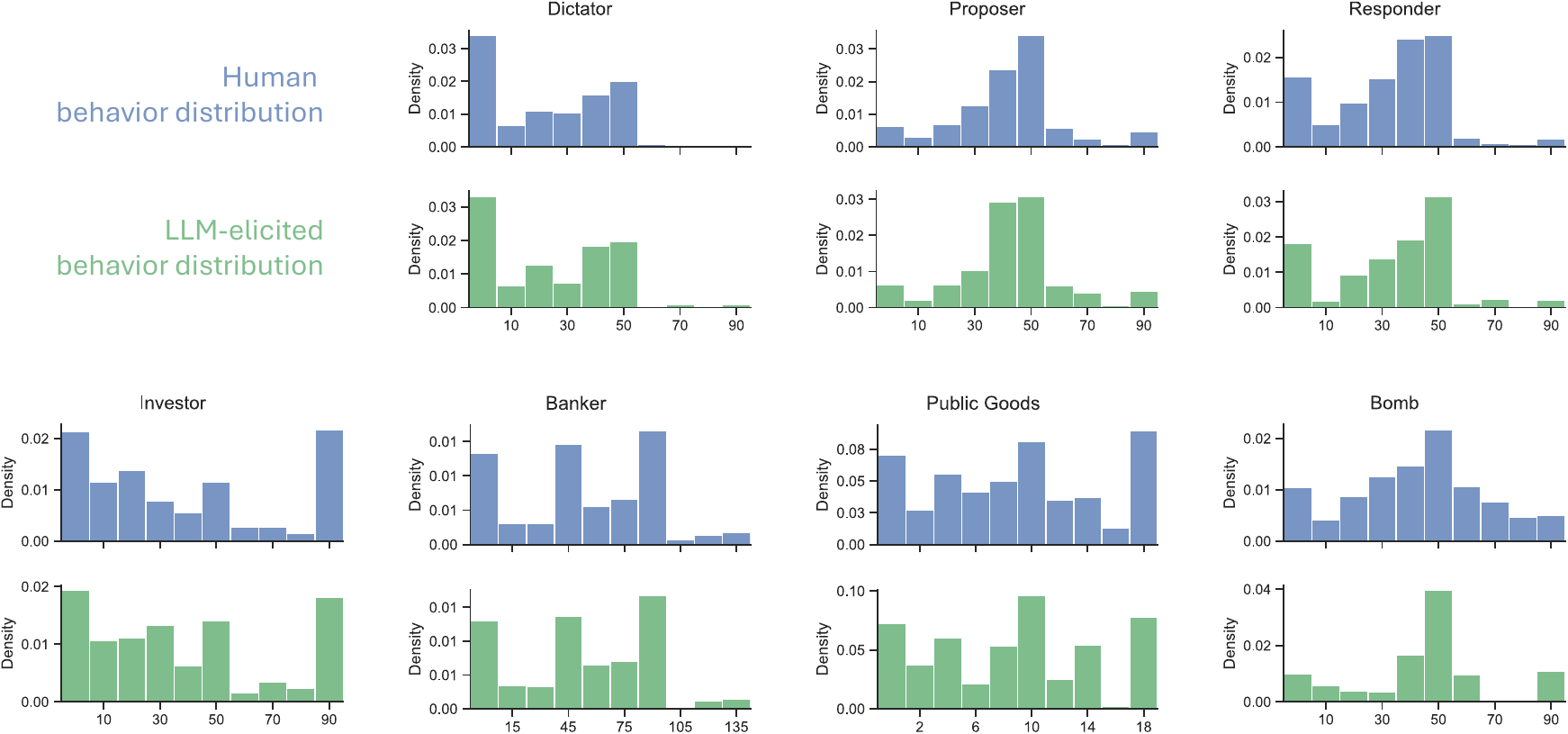}
        \caption{
        Behavior distributions observed from human player data (\textcolor{myblue}{\textbf{blue}} histograms) and LLM-elicited distributions (\textcolor{mygreen}{\textbf{green}} histograms). By utilizing a mixture of the deciphered behavioral codes as the system prompts, LLM behaviors can be effectively aligned with the behavioral distributions of a human population. 
        } 
        \label{fig:dist-hist}
    \end{subfigure}
    \vspace{-10pt}
    \caption{\textbf{Illustration of how LLM can decipher and elicit behavior distributions. }}
    \label{fig:dist-align}
\end{figure}

As illustrated in Figure \ref{fig:dist-illus}, an observed behavior distribution can be recreated from a mixture of behavioral codes. Algorithm \ref{alg:dist-align} outlines the procedure for determining the optimal mixture of codes to achieve distributional alignment with an observed distribution.
For the OptimizeVector function in our experiments, we used the Sequential Least Squares Programming (SLSQP) algorithm from the \texttt{scikit-learn} package. The optimization was performed with the linear constraint $\sum w_i=1$ and bounds $w_i\in[0,1]$. We set the termination tolerance to $10^{-6}$ and the regularization parameter $\alpha$ to 3.

After optimizing the weights ($\boldsymbol{w}$) over behavioral codes ($\mathcal{X}$) as described in Algorithm \ref{alg:dist-align}, the mixture of codes is evaluated by comparing the LLM-elicited behavior distribution with the observed human behavior distribution, as shown in Figure \ref{fig:dist-hist}. For this evaluation, the mixture generates 1000 samples by sampling a behavioral code according to the weights and eliciting a behavior.

The distributional alignment successfully fits the behavioral codes to a distribution of human play. Table~\ref{tab:dist-align} shows that the LLM-elicited distribution given the deciphered mixture of behavioral codes is indistinguishable from the human distribution in 5 out of 7 games, under a relaxed Kolmogorov-Smirnov (KS) test\footnote{In the relaxed Kolmogorov-Smirnov (KS) test, samples are binned with a width of 10 (the bin width for the Public Goods game is 2).}. 

Table \ref{tab:dist-align} also presents the distribution similarity. To provide context, two baselines are included:
\begin{enumerate}[label=(\arabic*)]
    \item An upper bound: 1000 random samples are drawn from the human-play records to compare against the full human behavior distribution.
    \item A lower bound: 1000 behavior choices are generated using the default system prompt (``You are a helpful assistant.'') without leveraging the behavioral codes.
\end{enumerate}
The evaluation results demonstrate that using a mixture of system prompts improves the alignment of LLM behavior distributions with human behavior distributions compared to the default system prompt, highlighting AI’s ability to decipher and then elicit behavior distributions.

\begin{table}%
\caption{\textbf{Distributional alignment performance.} 
Columns represent game scenarios, and rows indicate different behavior generation/sampling methods. 
The elicited/sampled behaviors are then compared with the ground-truth human distributions. 
The performance is evaluated using the Wasserstein metric, which measures the distance between the LLM-elicited behavior distributions (with or without the behavior code mixture) or a subsample human distribution and the observed human distribution for each game. A lower Wasserstein distance indicates a closer alignment.
``$^{\ddagger}$'' denotes distributions that are statistically indistinguishable from the observed human distribution under the Kolmogorov–Smirnov (KS) test ($p > 0.05$), while ``$^{\dagger}$'' indicates indistinguishability under a relaxed KS test, where behavior choices are rounded to multiples of 5 (multiples of 2 for Public Goods) before statistical testing.
The results suggest that using a mixture of behavioral codes improves the alignment of LLM behavior distributions with human behavior distributions compared to the default system prompt, highlighting AI’s ability to decipher behavior distributions.
}
\footnotesize	
\centering
\begin{tabular}{|l|ccccccc|}
\hline
& \textbf{Dictator} & \textbf{Proposer} & \textbf{Responder} & \textbf{Investor} & \textbf{Banker} & \makecell[b]{\textbf{Public}\\\textbf{Goods}} & \textbf{Bomb} \\
\hline
\makecell[l]{\textcolor{myblue}{\textbf{Human Sample}}\\\textcolor{myblue}{1,000 random sample}}             & 0.69$^{\ddagger}$                                  & 0.72$^{\ddagger}$                                  & 0.70$^{\ddagger}$                                   & 1.11$^{\ddagger}$                                  & 0.85$^{\ddagger}$                                & 0.21$^{\ddagger}$                                      & 0.90$^{\ddagger}$                              \\
\hline
\makecell[l]{\textcolor{myorange}{\textbf{LLM-default distribution}}\\\textcolor{myorange}{Default system prompt}}       & 25.58                                 & 11.35                                 & 33.99                                  & 28.23                                 & 20.00                               & 25.49                                     & 16.62                             \\
\hline
\makecell[l]{\textcolor{mygreen}{\textbf{LLM-elicited distribution}}\\\textcolor{mygreen}{A mixture of behavioral codes}} 
& 0.74$^{\ddagger}$ & 0.92$^{\ddagger}$ & 1.11$^{\dagger}$ & 2.30 & 1.52$^{\dagger}$ & 2.27$^{\dagger}$ & 4.78 \\
\hline
\end{tabular}
\label{tab:dist-align}
\end{table}

We also investigate how learned behavioral code mixtures can generalize across games. We elicit behaviors from one (target) game using behavioral codes from another (source) game, and compared outcomes with human-play records in the target game. The Wasserstein distances of the elicited behavior distribution and the human distribution are shown in Table \ref{tab:transfer}. In particular, we see that the behavioral codes obtained from one game can elicit behavior distributions that are closer to the human distribution than the default system prompt (no behavioral code) in most game settings, where the most aligned behaviors are still elicited from codes derived from the same game. 

\begin{table}%
\caption{\textbf{The performance of using the behavioral code mixture learned from one game (source) to another game (target).} 
Columns represent target game scenarios, and rows indicate the source games where the behavioral code mixtures are learned from. The ``Default System Prompt`` stands for using the default system prompt without leveraging the behavioral codes. The elicited behaviors are then compared with the ground-truth human distributions. 
The performance is evaluated using the Wasserstein metric, which measures the distance between the LLM-elicited behavior distributions (with or without the behavior code mixture) or a subsample human distribution and the observed human distribution for each game. A lower Wasserstein distance indicates a closer alignment.
}
\footnotesize 
\centering
\begin{tabular}{|l|ccccccc|}
\hline
& \textbf{Dictator} & \textbf{Proposer} & \textbf{Responder} & \textbf{Investor} & \textbf{Banker} & \makecell[b]{\textbf{Public}\\\textbf{Goods}} & \textbf{Bomb} \\
\hline
\textbf{Default System Prompt} & 25.58 & 11.95 & 32.53 & 42.78 & 20.35 & 24.00 & 26.99 \\
\hline
\textbf{Dictator} & 0.74 & 17.94 & 12.77 & 8.49 & 4.05 & 20.83 & 9.49 \\
\textbf{Proposer} & 22.41 & 0.92 & 9.37 & 15.29 & 19.86 & 8.11 & 10.10 \\
\textbf{Responder} & 11.89 & 10.63 & 1.11 & 25.11 & 14.73 & 13.78 & 11.33 \\
\textbf{Investor} & 17.99 & 9.38 & 14.86 & 2.30 & 25.09 & 8.53 & 22.41 \\
\textbf{Banker} & 5.67 & 17.57 & 18.73 & 13.37 & 1.52 & 24.88 & 12.19 \\
\textbf{Public Goods} & 15.16 & 7.28 & 15.04 & 9.60 & 20.70 & 2.27 & 13.35 \\
\textbf{Bomb} & 16.53 & 8.89 & 5.63 & 19.25 & 20.61 & 13.68 & 4.78 \\
\hline
\end{tabular}
\label{tab:transfer}
\end{table}

\section{Supplementary Text}

\subsection{Interpretability of Behavioral Codes}
\label{app:interpret}

To verify the interpretability of the learned behavioral codes, we perform a textual analysis. Specifically, we first preprocess the behavioral codes using the \texttt{spaCy} library with the \texttt{en\_core\_web\_sm} model. This preprocessing involved tokenization, lemmatization, and the removal of stop words (using \texttt{.is\_stop}) and non-alphabetic tokens (using \textbf{.is\_alpha}).
Each preprocessed behavioral code is then represented as a bag of keywords, and the TF-IDF importance score of each keyword within the code is calculated. Table \ref{tab:keywords} presents the top 50 keywords (those with highest aggregated TF-IDF scores) from the behavioral codes for each game scenario.

\begin{table}%
\caption{\textbf{Top 50 keywords in behavioral codes for each game scenario. } 
Each column represent a game scenario, and keywords are list in rows, sorted by TF-IDF word importance. }
\scriptsize
\centering
\begin{tabular}{|r|l|l|l|l|l|l|l|}
\hline
& \textbf{Dictator} & \textbf{Proposer} & \textbf{Responder} & \textbf{Investor} & \textbf{Banker} & \textbf{Public Goods} & \textbf{Bomb} \\
\hline
1 & decision & proposal & decision & decision & investor & group & risk \\
2 & fairness & decision & fairness & risk & profit & contribution & decision \\
3 & strategic & player & proposal & investment & decision & benefit & reward \\
4 & maker & strategic & ensure & potential & ensure & decision & potential \\
5 & aim & ensure & outcome & return & balance & personal & high \\
6 & balance & party & benefit & aim & term & collective & maximize \\
7 & outcome & aim & reflect & balance & strategic & aim & balance \\
8 & benefit & outcome & strategic & strategic & trust & maximize & gain \\
9 & ensure & offer & maker & reflect & maximize & balance & aim \\
10 & choice & fairness & offer & investor & aim & contribute & outcome \\
11 & reflect & maximize & aim & maximize & future & individual & choice \\
12 & generosity & acceptance & fair & approach & benefit & strategic & probability \\
13 & party & benefit & value & choice & gain & outcome & maker \\
14 & maximize & accept & gain & growth & long & payoff & point \\
15 & fair & balance & maximize & make & relationship & ensure & strategic \\
16 & value & fair & balance & ensure & maintain & gain & approach \\
17 & make & likely & accept & high & return & overall & minimize \\
18 & consider & consider & prioritize & outcome & foster & make & make \\
19 & maintain & propose & consider & gain & fairness & return & carefully \\
20 & prioritize & maker & high & conservative & fair & maker & achieve \\
21 & self & mutual & secure & maker & banker & high & optimal \\
22 & resource & negotiator & evaluate & cautious & investment & resource & strategist \\
23 & advantage & make & rational & trust & encourage & consider & loss \\
24 & thoughtful & goal & reasonable & maintain & party & success & calculate \\
25 & slightly & generosity & share & moderate & reflect & prioritize & strategy \\
26 & generous & likelihood & equitable & reward & cooperation & reflect & optimize \\
27 & create & focus & achieve & carefully & player & optimize & safety \\
28 & gain & understand & choice & resource & maker & participant & scenario \\
29 & positive & generous & substantial & achieve & value & cooperative & avoid \\
30 & sense & reflect & focus & seek & mutual & approach & ensure \\
31 & goodwill & gain & receive & balanced & immediate & term & prioritize \\
32 & strive & cooperation & strong & strategy & outcome & focus & excel \\
33 & favor & create & scenario & prioritize & consider & carefully & reflect \\
34 & approach & value & carefully & optimize & ongoing & game & possible \\
35 & focus & prioritize & self & consider & focus & long & cautious \\
36 & reasonable & foster & possible & prudent & interaction & potential & conservative \\
37 & considerate & positive & standard & significant & continue & strategy & significant \\
38 & scenario & high & sense & invest & collaboration & optimal & consider \\
39 & equitable & highly & pragmatic & minimize & prioritize & goal & involve \\
40 & optimize & beneficial & negotiation & calculate & optimize & scenario & focus \\
41 & foster & understanding & acceptable & prefer & importance & possible & evaluate \\
42 & involve & reasonable & negotiator & opportunity & positive & effort & favor \\
43 & seek & importance & personal & thoughtful & goal & enhance & goal \\
44 & share & empathetic & goal & careful & understand & cooperation & level \\
45 & balanced & optimize & significant & level & self & collaborative & choose \\
46 & personal & involve & threshold & substantial & secure & achieve & option \\
47 & retain & achieve & make & calculated & reward & balanced & success \\
48 & achieve & self & meet & player & strive & foster & calculated \\
49 & understand & agreement & maintain & caution & feel & slightly & thinker \\
50 & offer & perceive & advantageous & small & win & cautious & bold \\
\hline
\end{tabular}
\label{tab:keywords}
\end{table}

To understand which keywords influence the behaviors, as deciphered by LLMs, we perform a linear regression analysis. Treating each behavioral code as an observation, we generate 10 behavior choices with this code and predict the mean behavior based on the occurrence of keywords. An ordinary least squares (OLS) linear regression model is applied using binary keyword occurrence feature vectors with 50 dimensions. Table \ref{tab:ols-full} presents the complete linear regression results, showing the impact of each keyword in the game scenarios. Figure \ref{fig:lr-ols} 
highlights the top 10 keywords in the linear regression analysis. 
To alleviate the influence of multicollinearity, we also report the LASSO regression results in Table \ref{tab:ols-full-lasso}. A comparison of the OLS and LASSO results reveals strong alignment among the top keywords (those with high absolute LR coefficients), suggesting the robustness of the observed correlations.

\begin{table}%
\caption{\textbf{Linear regression analysis of behaviors based on keywords in the behavioral codes.} 
Specifically, each behavioral code is transformed into a binary keyword-occurrence feature vector of 50 dimensions to predict the mean behavior that can be elicited by that code. This prediction is performed using an ordinary least squares (OLS) linear regression model. 
The results demonstrate that keyword occurrences in behavioral codes effectively predict the behaviors of LLMs, offering interpretability of the behavioral codes.}
\centering
\begin{tabular}{c}
    \includegraphics[width=\linewidth]{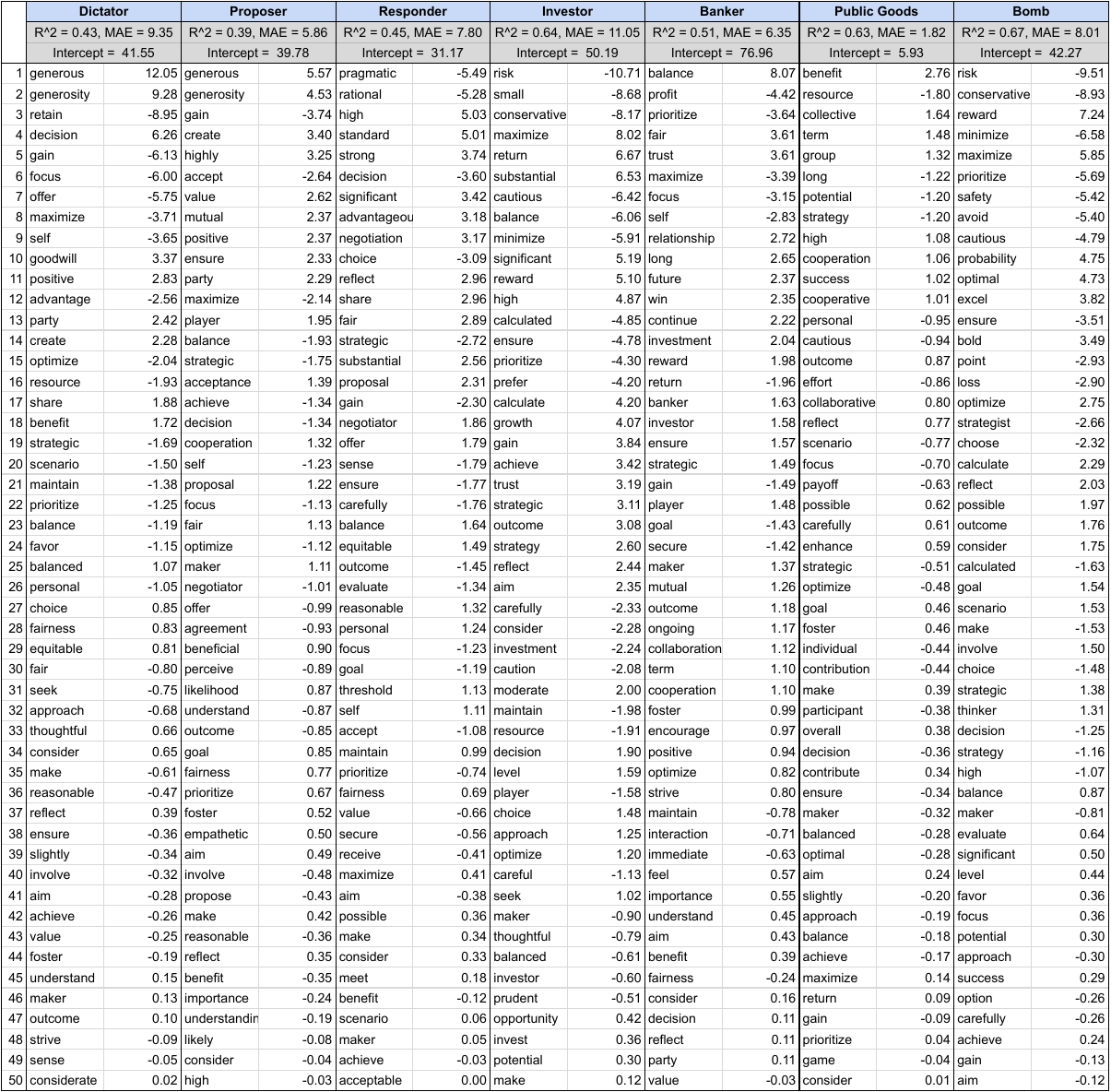}
\end{tabular}
\label{tab:ols-full}
\end{table}

\begin{table}%
\caption{\textbf{Linear regression analysis of behaviors based on keywords in the behavioral codes.} 
Specifically, each behavioral code is transformed into a binary keyword-occurrence feature vector of 50 dimensions to predict the mean behavior that can be elicited by that code. This prediction is performed using a LASSO linear regression model with the regularization coefficient ($\alpha=0.3$). 
The results demonstrate that keyword occurrences in behavioral codes effectively predict the behaviors of LLMs, offering interpretability of the behavioral codes.}
\centering
\begin{tabular}{c}
    \includegraphics[width=\linewidth]{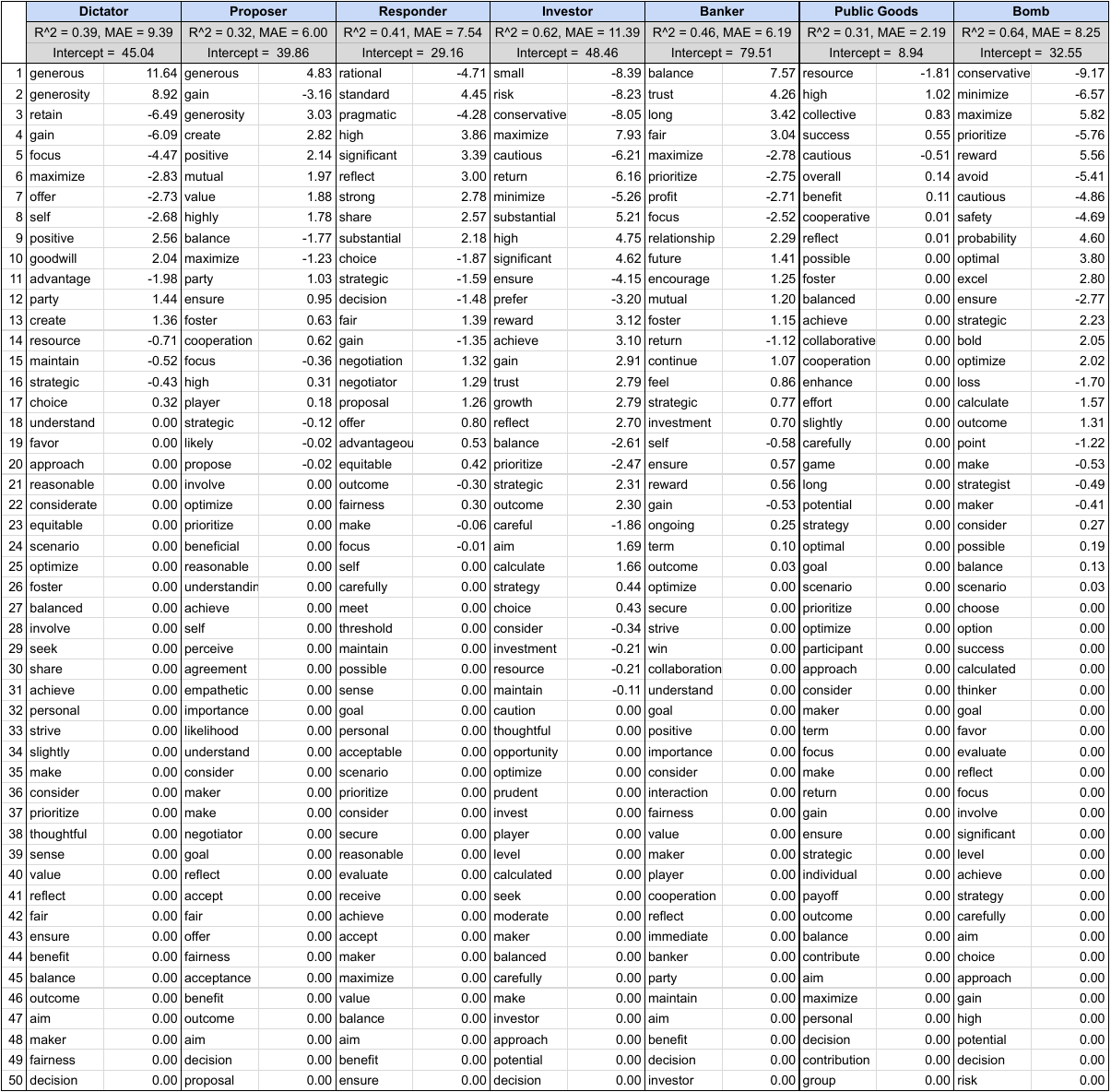}
\end{tabular}
\label{tab:ols-full-lasso}
\end{table}

\begin{figure}
    \centering
    \includegraphics[height=22cm]{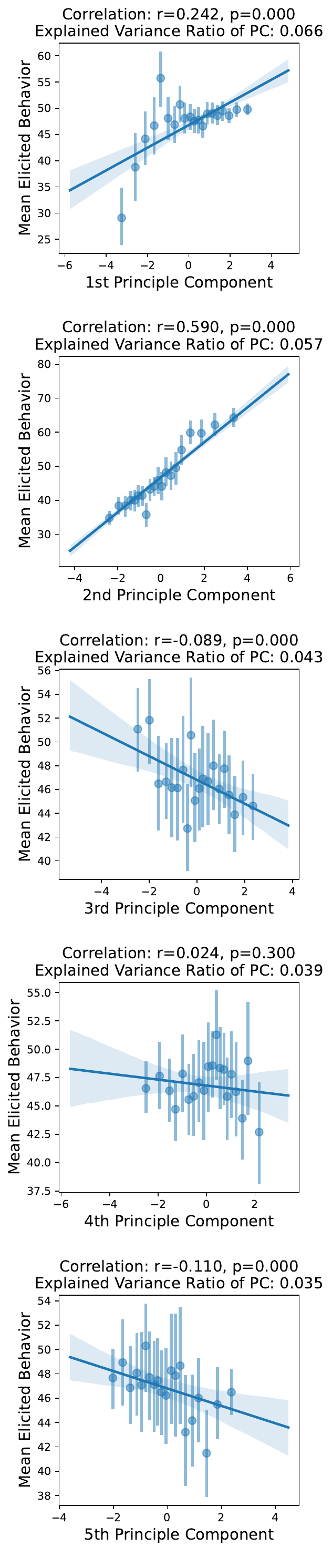}
    \hspace{30pt}
    \includegraphics[height=22cm]{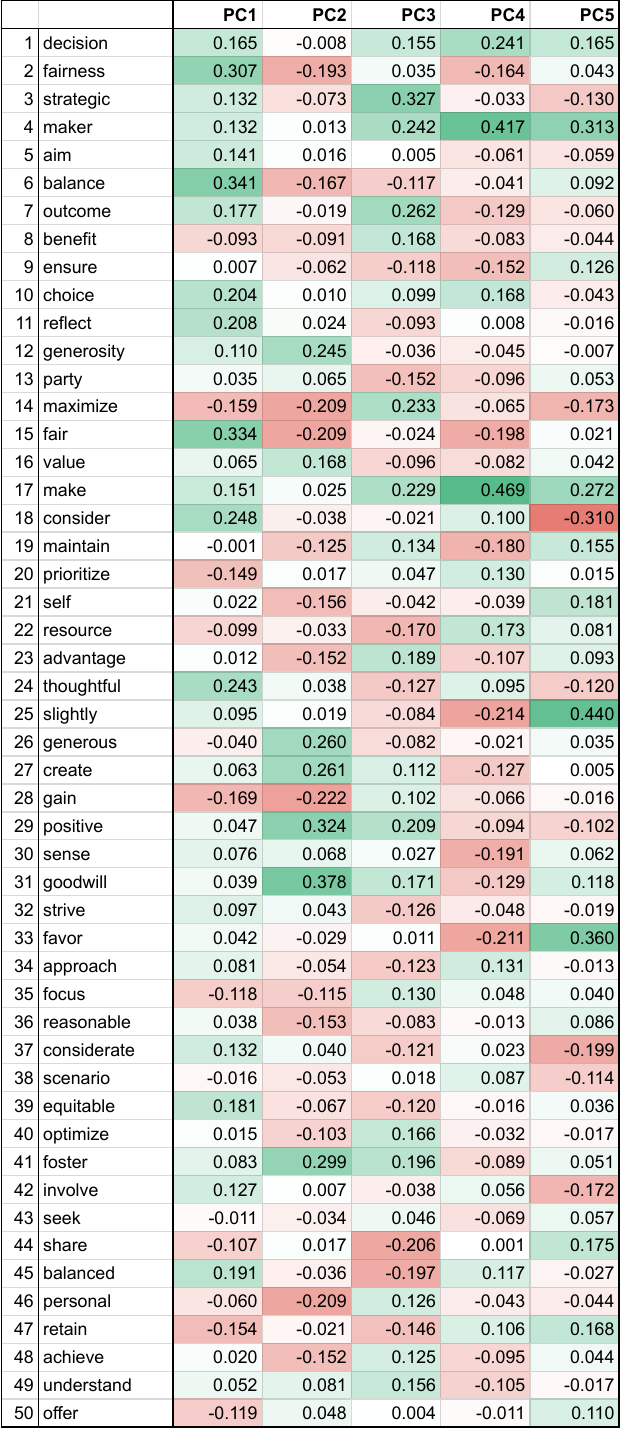}
    \caption{
    \textbf{Correlations between the elicited behaviors and behavioral codes' PCs (left) and the loading vectors of PCs (right) for the Dictator game. }
    }
    \label{fig:pca-five-dictator}
\end{figure}

\begin{figure}
    \centering
    \includegraphics[height=22cm]{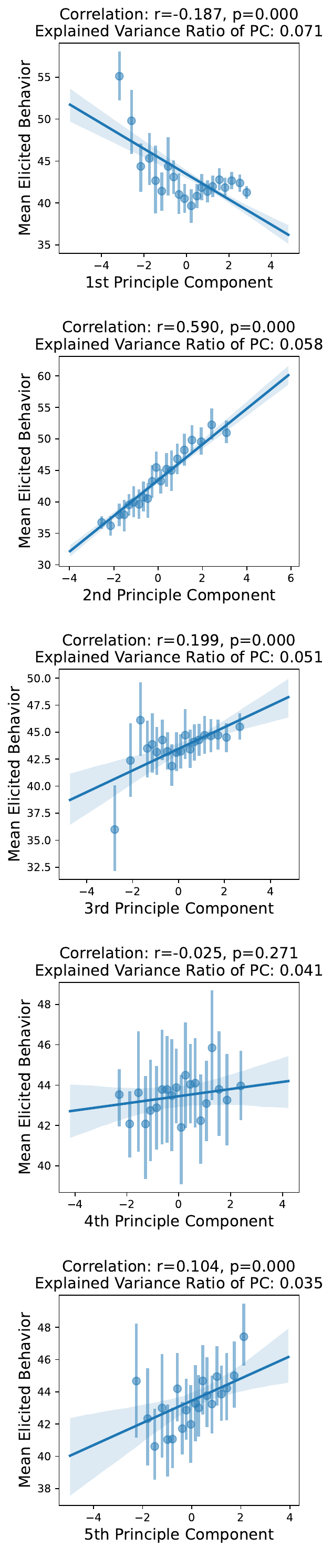}
    \hspace{30pt}
    \includegraphics[height=22cm]{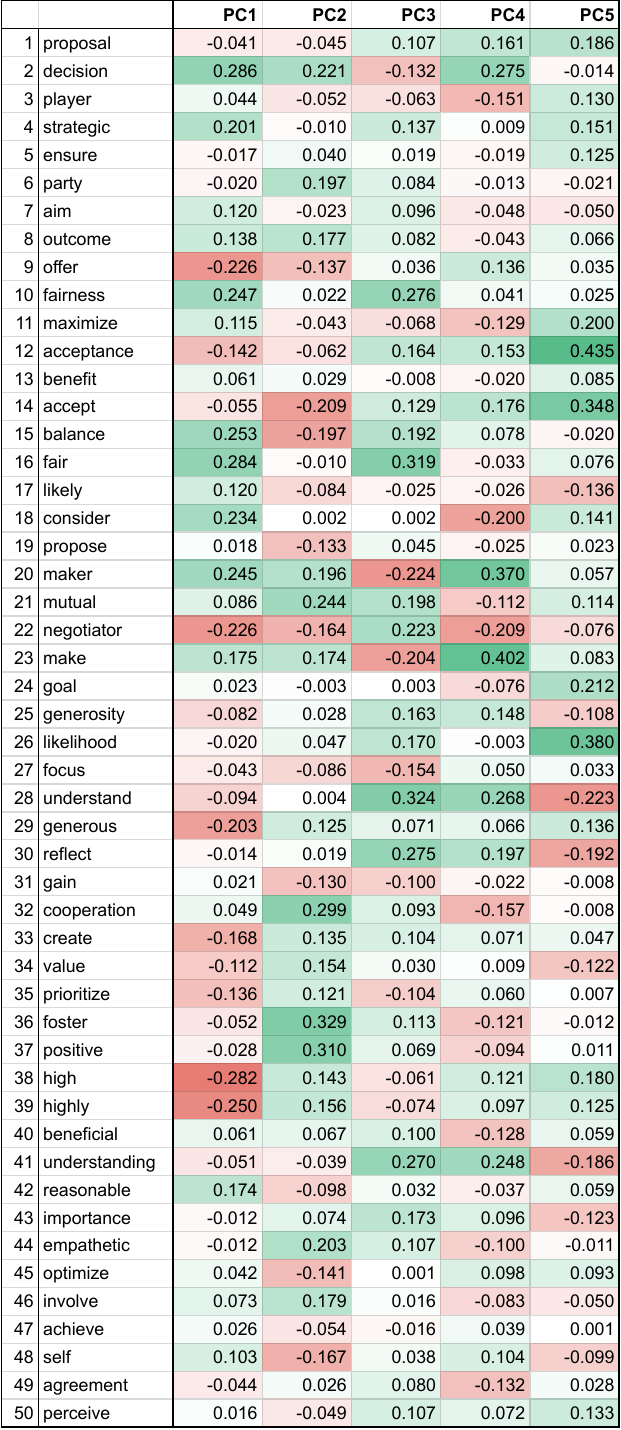}
    \caption{
    \textbf{Correlations between the elicited behaviors and behavioral codes' PCs (left) and the loading vectors of PCs (right) for the Proposer game. }
    }
    \label{fig:pca-five-proposer}
\end{figure}

\begin{figure}
    \centering
    \includegraphics[height=22cm]{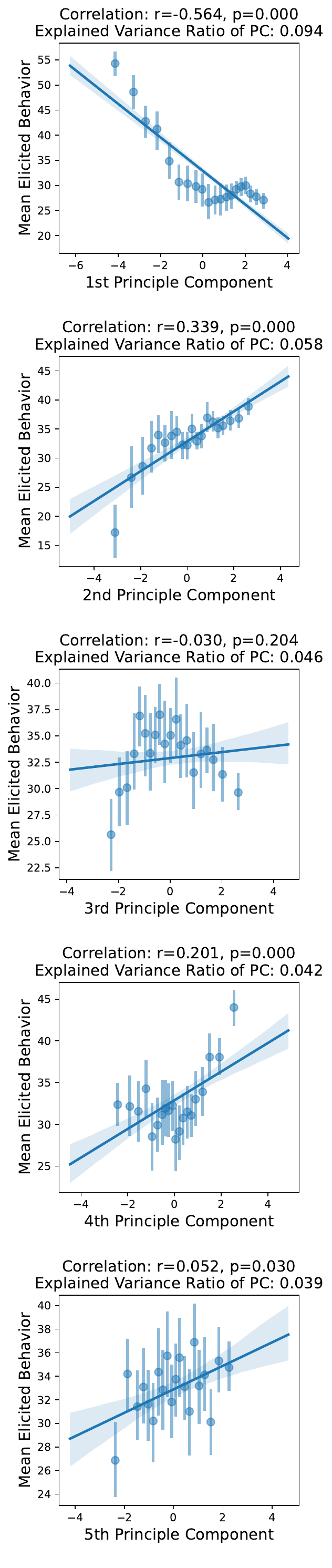}
    \hspace{30pt}
    \includegraphics[height=22cm]{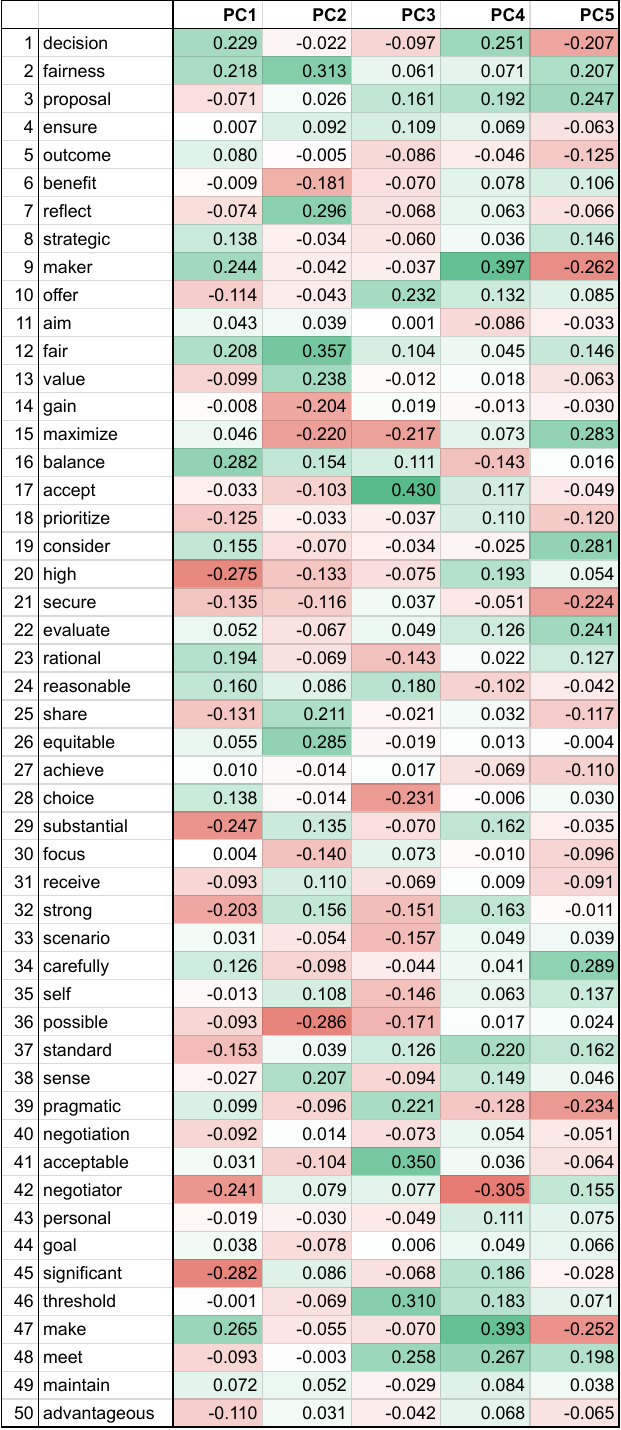}
    \caption{
    \textbf{Correlations between the elicited behaviors and behavioral codes' PCs (left) and the loading vectors of PCs (right) for the Responder game. }
    }
    \label{fig:pca-five-responder}
\end{figure}

\begin{figure}
    \centering
    \includegraphics[height=22cm]{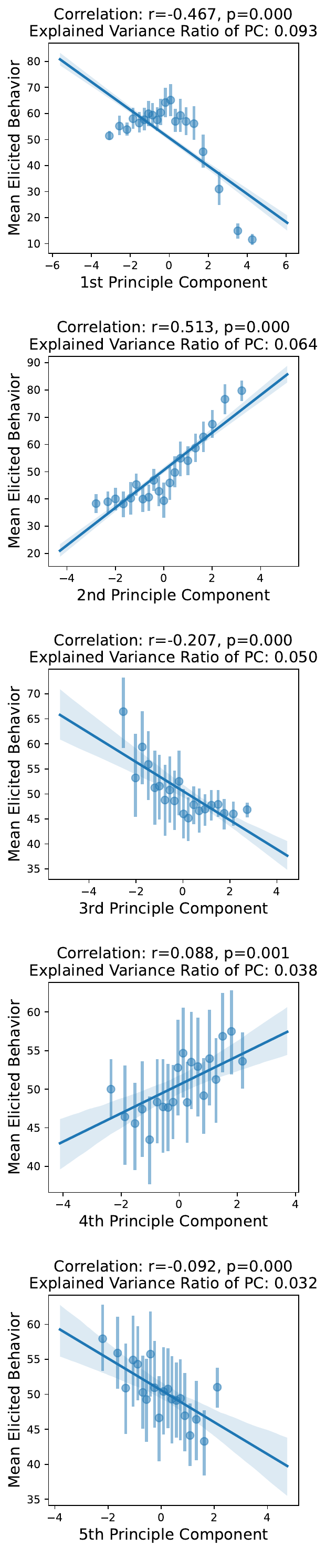}
    \hspace{30pt}
    \includegraphics[height=22cm]{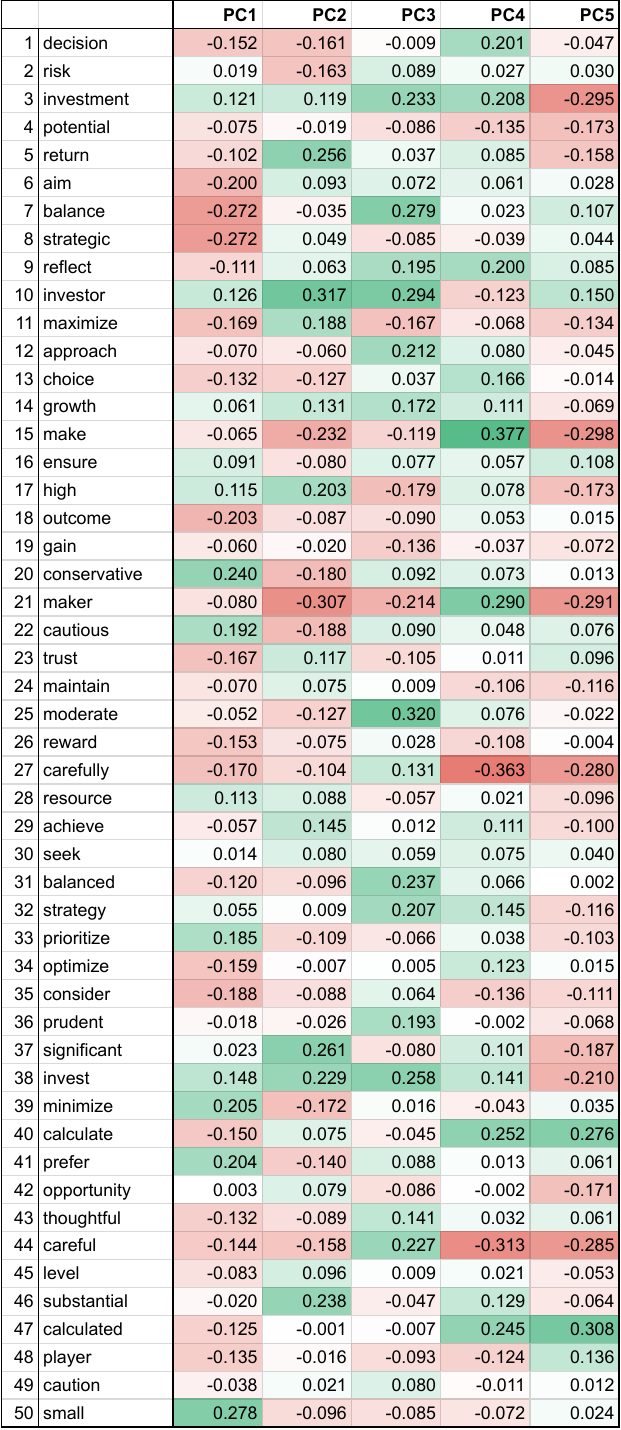}
    \caption{
    \textbf{Correlations between the elicited behaviors and behavioral codes' PCs (left) and the loading vectors of PCs (right) for the Investor game. }
    }
    \label{fig:pca-five-investor}
\end{figure}

\begin{figure}
    \centering
    \includegraphics[height=22cm]{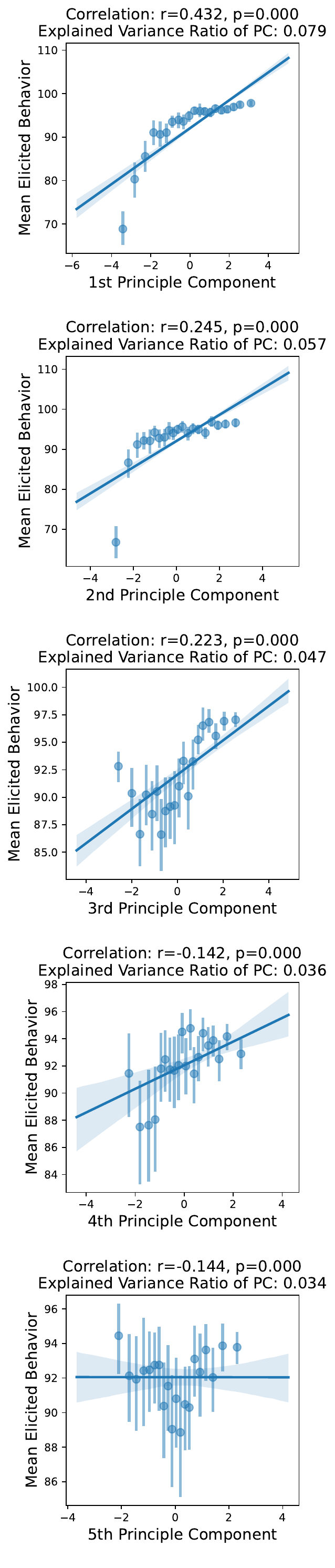}
    \hspace{30pt}
    \includegraphics[height=22cm]{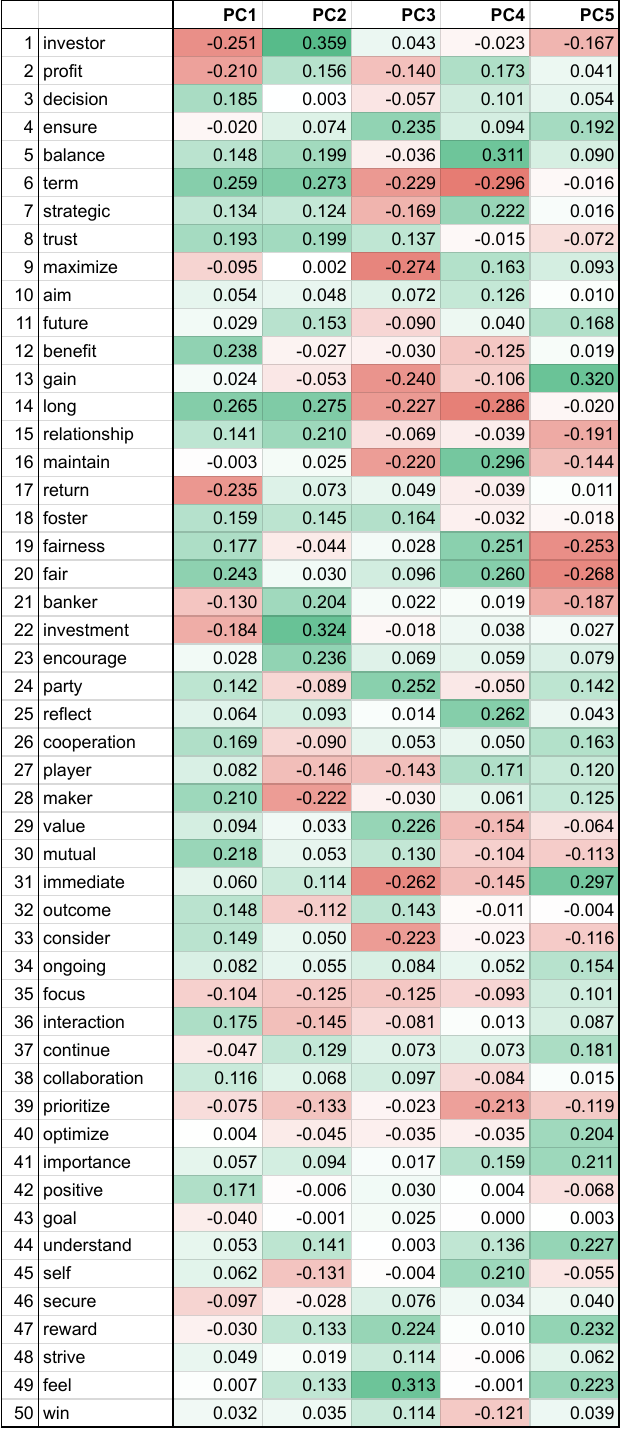}
    \caption{
    \textbf{Correlations between the elicited behaviors and behavioral codes' PCs (left) and the loading vectors of PCs (right) for the Banker game. }
    }
    \label{fig:pca-five-banker}
\end{figure}

\begin{figure}
    \centering
    \includegraphics[height=22cm]{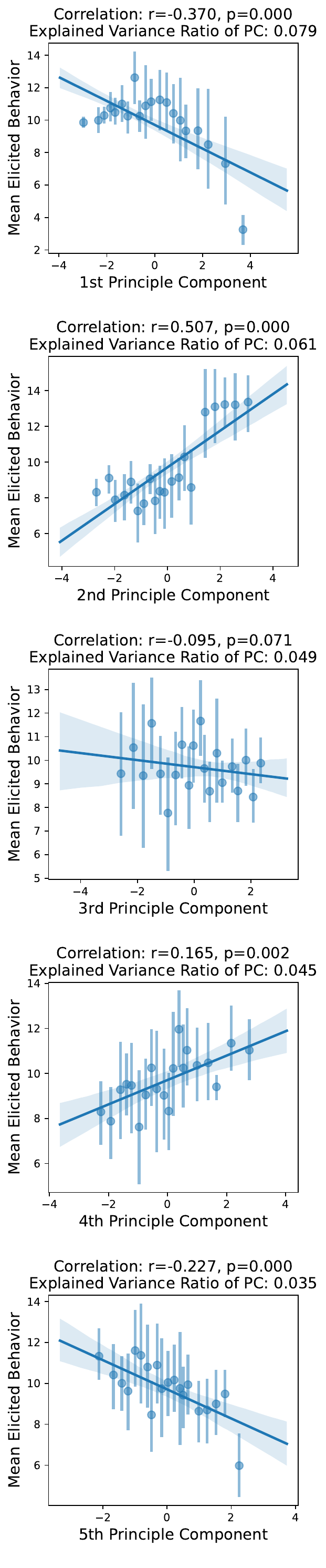}
    \hspace{30pt}
    \includegraphics[height=22cm]{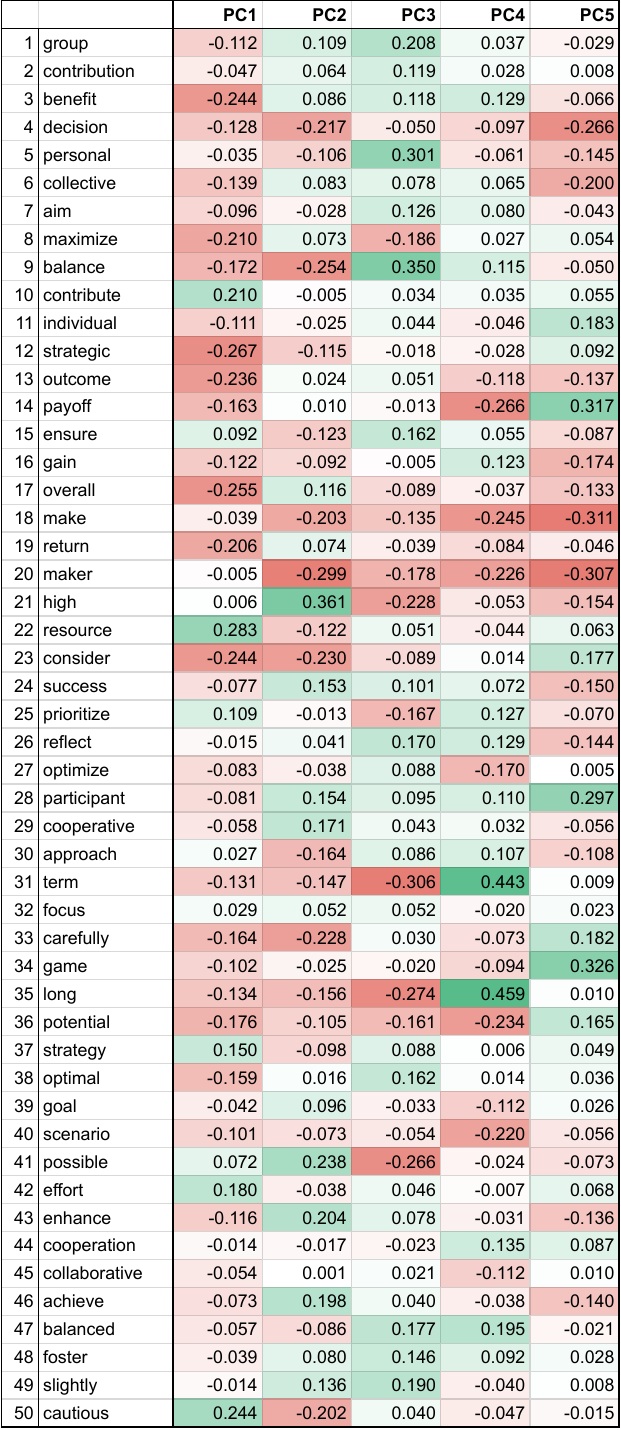}
    \caption{
    \textbf{Correlations between the elicited behaviors and behavioral codes' PCs (left) and the loading vectors of PCs (right) for the Public Goods game. }
    }
    \label{fig:pca-five-pg}
\end{figure}

\begin{figure}
    \centering
    \includegraphics[height=22cm]{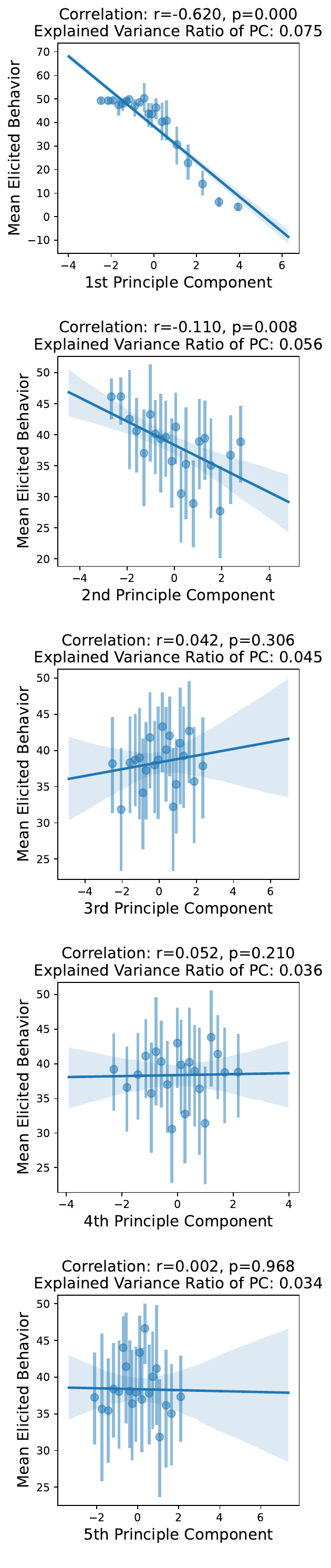}
    \hspace{30pt}
    \includegraphics[height=22cm]{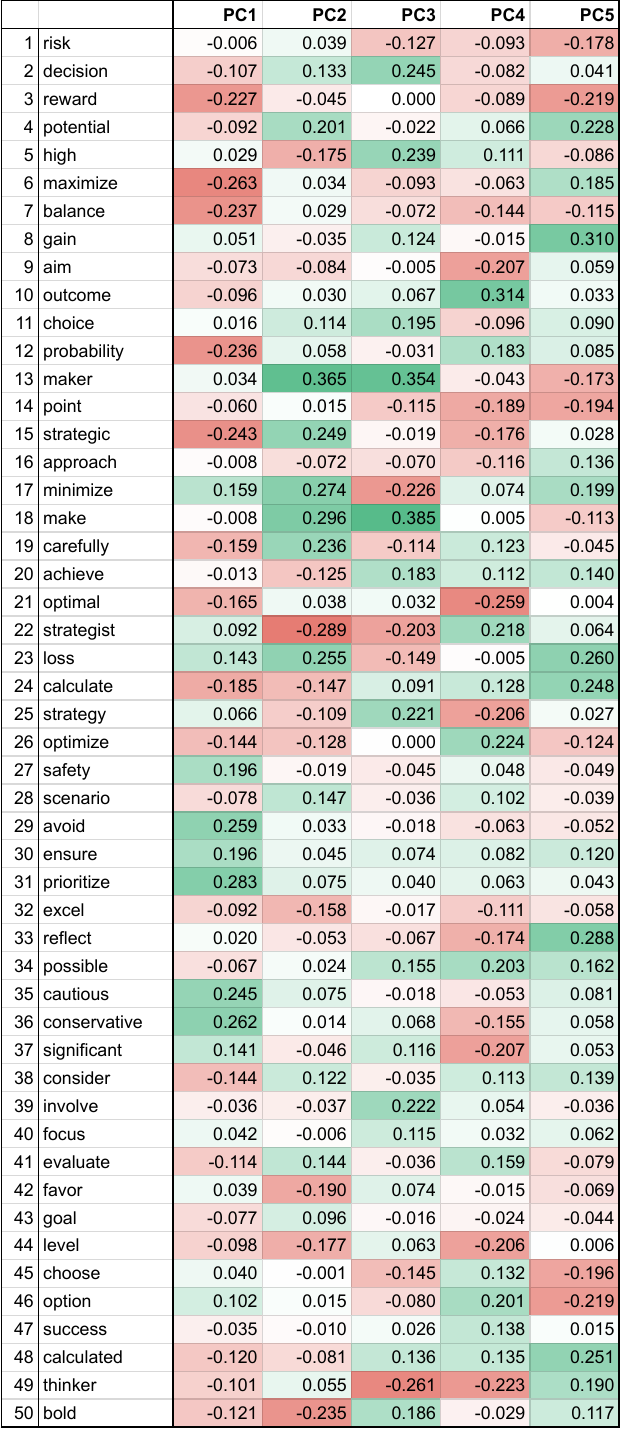}
    \caption{
    \textbf{Correlations between the elicited behaviors and behavioral codes' PCs (left) and the loading vectors of PCs (right) for the Bomb game. }
    }
    \label{fig:pca-five-bomb}
\end{figure}

\begin{figure}%
    \centering
    \begin{subfigure}[t]{0.24\textwidth}
        \hspace*{\fill} %
    \end{subfigure}
    \hfill
    \begin{subfigure}[t]{0.24\linewidth}
        \includegraphics[width=\linewidth]{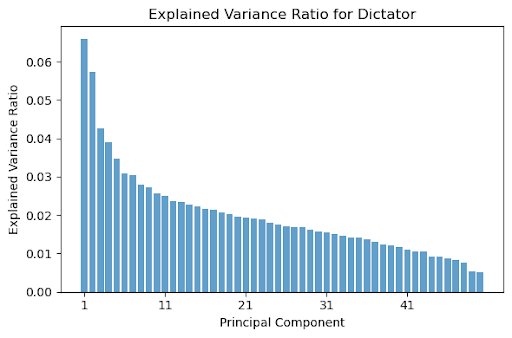}
    \end{subfigure}
    \hfill
    \begin{subfigure}[t]{0.24\linewidth}
        \includegraphics[width=\linewidth]{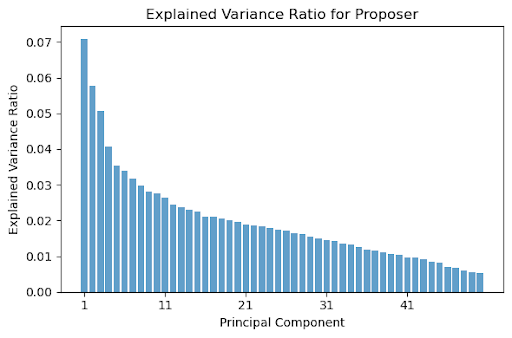}
    \end{subfigure}
    \hfill
    \begin{subfigure}[t]{0.24\linewidth}
        \includegraphics[width=\linewidth]{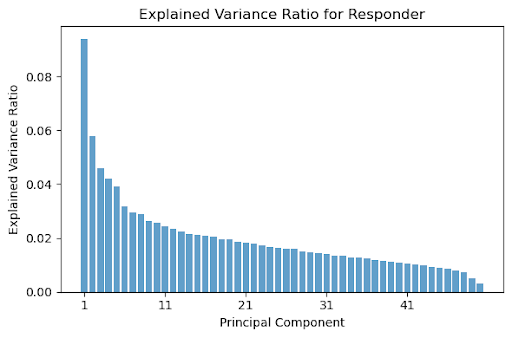}
    \end{subfigure}
    \begin{subfigure}[t]{0.24\linewidth}
        \includegraphics[width=\linewidth]{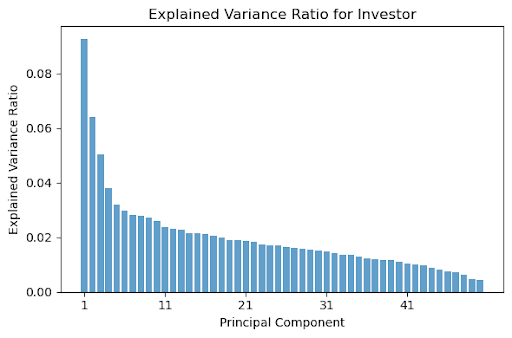}
    \end{subfigure}
    \hfill
    \begin{subfigure}[t]{0.24\linewidth}
        \includegraphics[width=\linewidth]{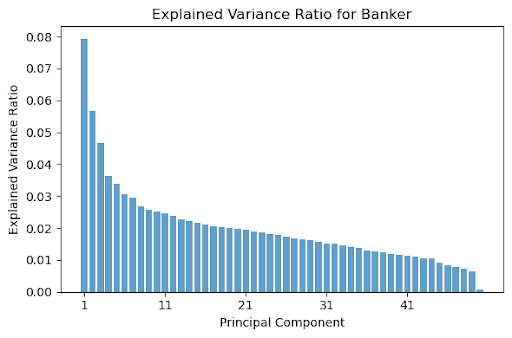}
    \end{subfigure}
    \hfill
    \begin{subfigure}[t]{0.24\linewidth}
        \includegraphics[width=\linewidth]{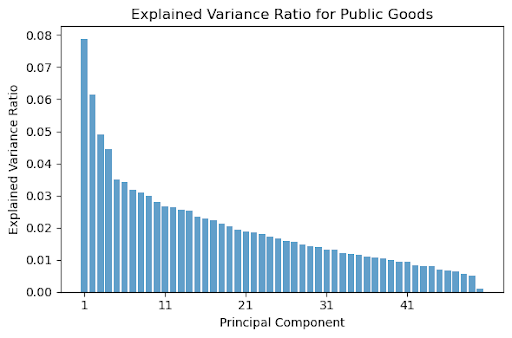}
    \end{subfigure}
    \hfill
    \begin{subfigure}[t]{0.24\linewidth}
        \includegraphics[width=\linewidth]{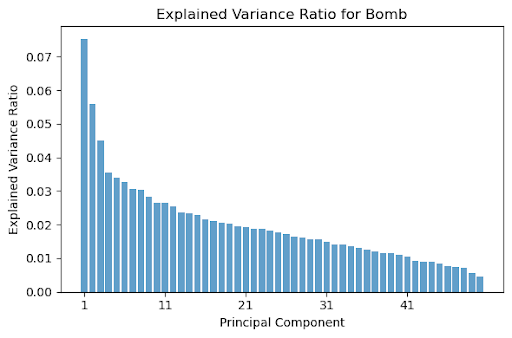}
    \end{subfigure}
    \caption{\textbf{Explained variance ratio for each principle component (PC). } The x-axis indicates the 50 PCs in order, and the y-axis is the explained variance ratio. }
    \label{fig:pca-vars}
\end{figure}

We further explore the deciphered information contained in the obtained behavioral codes by analyzing the correlations between the elicited behaviors and the principal components (PCs) of the behavioral codes. The PCs are derived from the 50-dimensional binary keyword vectors representing the codes for each game. Each code is used to elicit a behavior 10 times, and the Spearman’s correlation between the mean elicited behavior and the PC value is computed. Figures \ref{fig:pca-five-dictator}-\ref{fig:pca-five-bomb} show the first five PCs and their loading vectors, where moderate to strong correlations are observed, demonstrating the effectiveness of behavioral codes in deciphering behaviors.
Figure \ref{fig:pca-vars} show the decay of the explained variance ratios for the PCs of keywords in each game. The ratios decay smoothly, aligning with the common trend observed when applying SVD to sparse document-keyword vectors \cite{min2010decomposing}.

\begin{figure}%
    \centering
    \includegraphics[width=.95\linewidth]{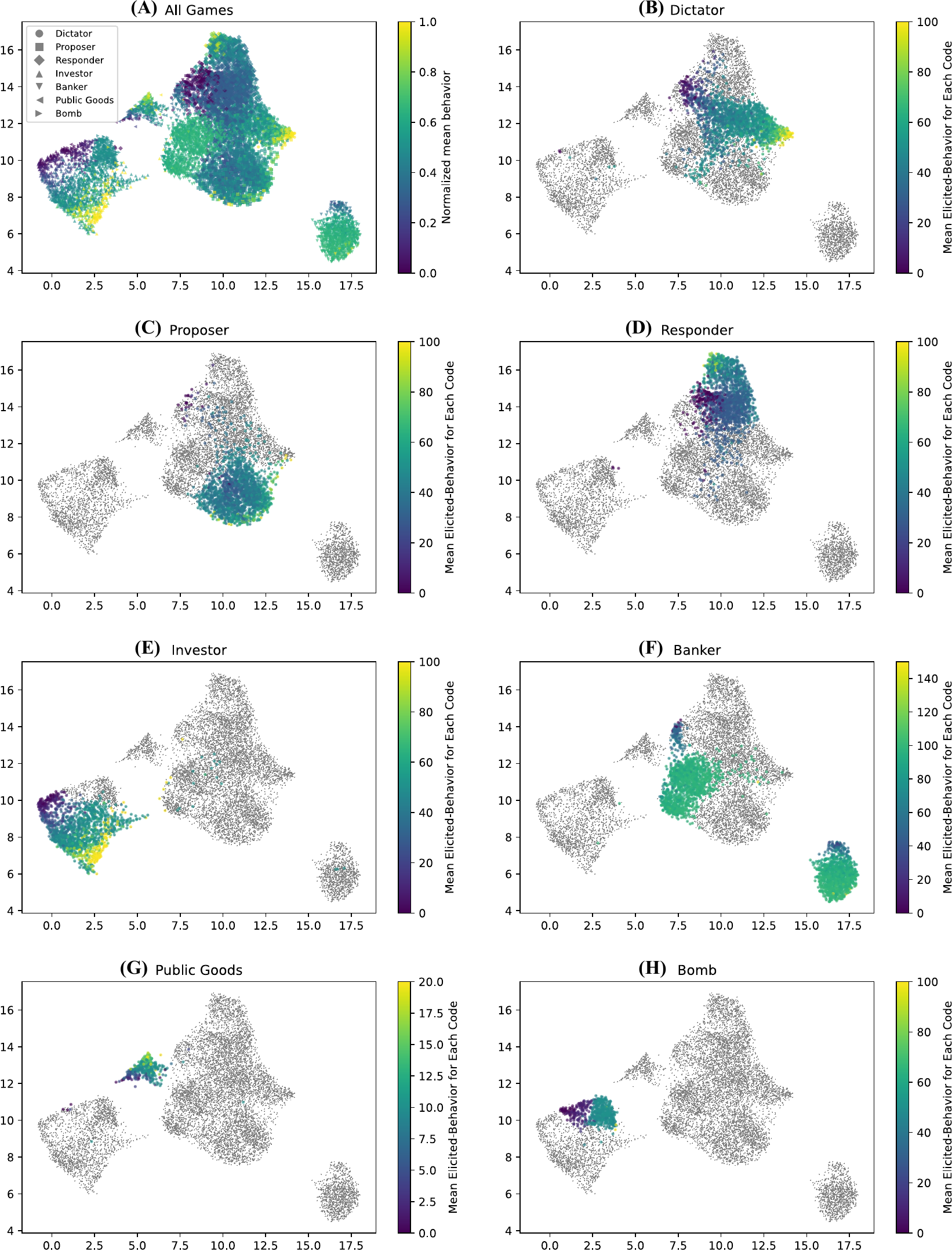}
    \caption{\textbf{The 2D projection of the behavioral codes across all games. }
    For each behavioral code, the mean behavior that is elicited is calculated (and normalized to the range $[0, 1]$). Corresponding colors are applied to the codes based on these values. 
    This visualization highlights the continuity of the behavioral codes with respect to their elicited behaviors. 
    }
    \label{fig:vis-values}
\end{figure}

The interpretability of behavioral codes is also consistent and generalizable. As illustrated in Fig \ref{fig:vis-values}, the smoothness of behaviors regarding the behavioral code semantics can be observed: Behavioral codes that are close in the semantic space tend to elicit similar behaviors, highlighting the structured and predictable nature of their influence on behavior generation.

\newpage
\subsection{Game Characteristics and Correlations}

Behavioral codes offer new insights into games. From the keywords used in game scenarios (Table \ref{tab:keywords}), we observe notable inter-game relationships. To further dissect these correlations, we analyze behavioral codes in a semantic space. Specifically, each behavioral code is embedded into a 1536-dimensional space using the embedding model OpenAI Ada 3 (\texttt{text-embedding-3-large}). These high-dimensional vectors are projected onto a 2D low-dimensional map using UMAP. The resulting projection map is displayed in Figure \ref{fig:vis-game}, 
with a heatmap highlighting semantic similarities between games. The similarities are computed as the aggregated cosine similarity of the high-dimensional embeddings.

\subsection{Behavioral Signatures for Populations}
\label{app:pop-signature}

By aligning the behavior distribution with a target population, as detailed in Sec. \ref{app:method-dist-align}, LLMs effectively decipher the signatures of this population into a mixture of behavioral codes.

For the MobLab player population, the weighted behavioral codes are highlighted on the 2D projection map shown in Figure \ref{fig:vis-pop}. 
Semantic clusters on the map are summarized into keywords using ChatGPT with the prompt below. The projection reveals that the weighted codes are unevenly distributed across the space, offering detailed and nuanced insights into the characteristics of the player population.

\begin{framed}
Here are 23 clusters of behavioral codes in the provided file. A behavioral code is a piece of text that induces behaviors with a natural language description of motivation or context. Please summarize these behavioral code clusters in to short titles (e.g., within four words). 

The summarized cluster titles should be: (1) understandable, (2) semantically relevant to the behavioral codes inside the clusters, (3) well covering the whole cluster, (4) distinguishable from other clusters, (5) generate a generalizable cluster title and avoid including any information specific to particular game instructions (will be given below).

Before output, please go back and forth review and refine the summarized cluster titles so that they meet the requirements. Please output the cluster names in lines, followed by the summarized titles (e.g., ``cluster 8AE92B-496: ...''). 

Game instructions:
\{game\_instructions\}
\end{framed}

\begin{figure}
    \centering
    \includegraphics[width=\linewidth]{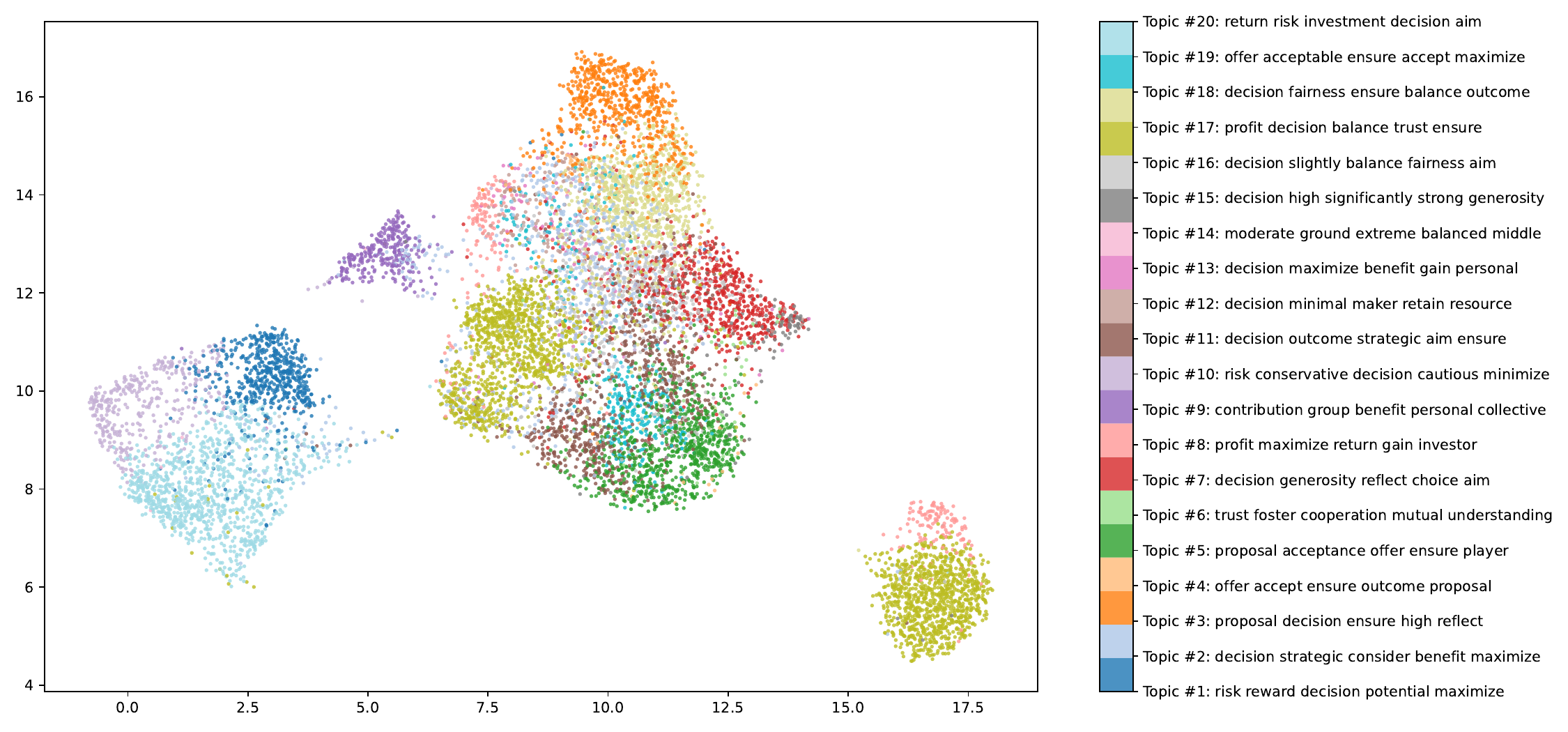}
    \caption{
    \textbf{Topic modeling results of the behavioral codes shown in the 2D projection map. }
    We apply the LDA topic modeling algorithm to the behavioral codes, each represented as a 50-dimensional binary keyword vector. Behavioral codes are colored by their dominant topic. Topics are summarized by their most correlated keywords.
    }
    \label{fig:topic-modeling}
\end{figure}

Figure \ref{fig:topic-modeling} presents an alternative annotation for the layout, derived through topic modeling. We apply the LDA topic modeling algorithm to the behavioral codes, with each code represented as a 50-dimensional binary keyword vector. A high alignment is observed between the semantic clusters in Figure \ref{fig:vis-pop} and the topic modeling results, yielding a Normalized Mutual Information (NMI) score of 0.594 between the clusters and the dominant topics of the codes.

Our topic modeling approach employed the Latent Dirichlet Allocation (LDA) class within the \texttt{scikit-learn} package. We set the number of topics (\texttt{n\_components}) to 20 and used a random state seed of 42 for reproducibility. Prior to modeling, behavioral codes were preprocessed using the \texttt{spaCy} library (specifically, the \texttt{en\_core\_web\_sm} model). This preprocessing involved tokenization, lemmatization, and the removal of stop words (using \texttt{.is\_stop}) and non-alphabetic tokens (using \texttt{.is\_alpha}). Subsequently, these processed codes were vectorized using scikit-learn's CountVectorizer, configured to remove 'english' stop words and limit features to a maximum of 1000.

\begin{figure}%
    \centering
    \includegraphics[width=\linewidth]{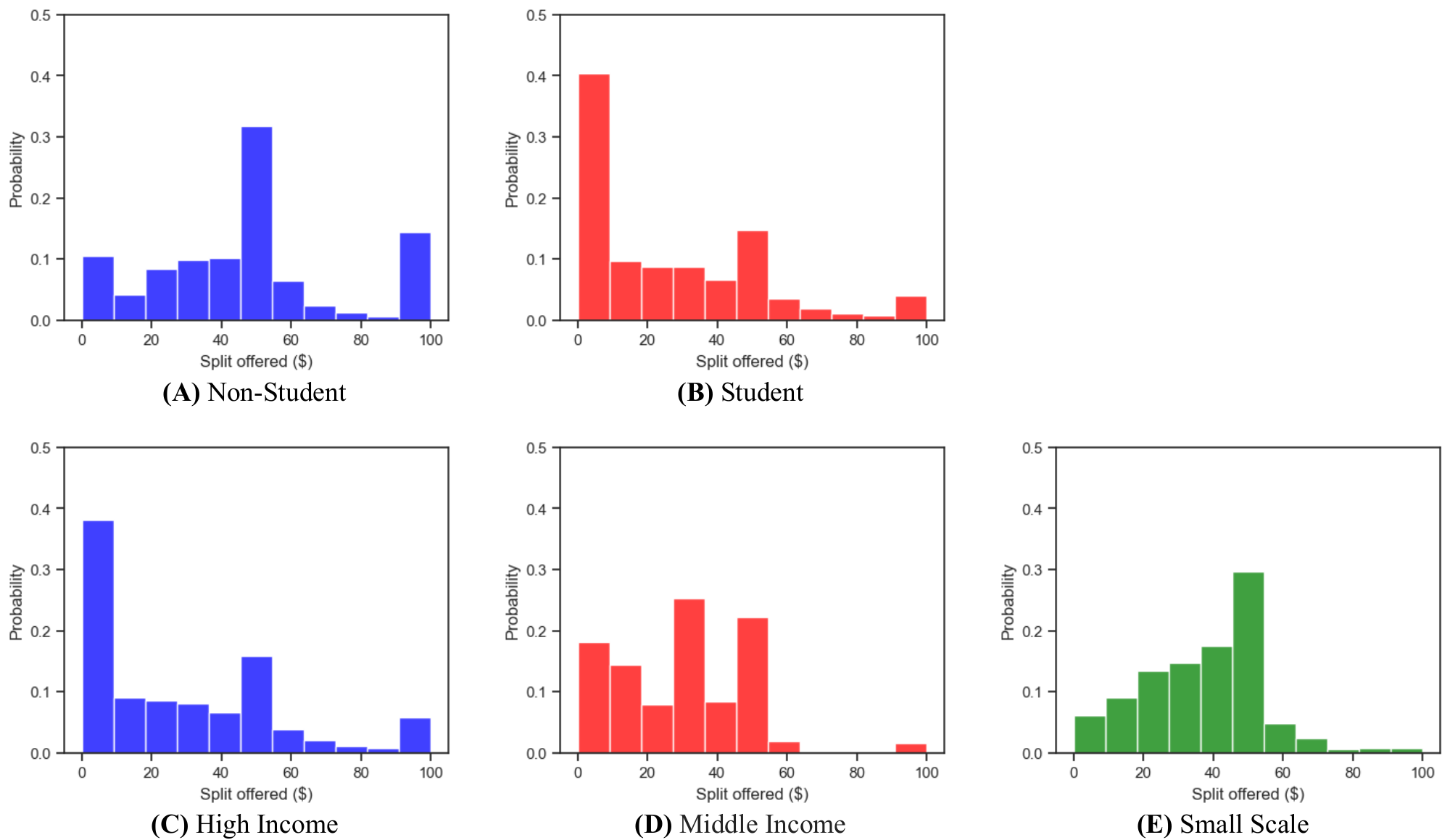}
    \caption{\textbf{Behavior distribution of different populations under the Dictator game.}}
    \label{fig:vis-pop-hist}
\end{figure}

Behavioral codes can also serve as a valuable tool for identifying distinct decision-making patterns across different populations. 
In a meta-study of the Dictator Game \cite{engel2011dictator}\footnote{Data from \cite{engel2011dictator}: \url{https://osf.io/xc73h/}, retrieved on March 14, 2025.}, various populations were analyzed, revealing distinct behavioral distributions.
Figure \ref{fig:vis-pop-hist} displays the behavior distributions of various populations in the Dictator game, including comparisons between Students and Non-students populations, as well as populations from diverse societal backgrounds. Figure \ref{fig:vis-pop-map} 
highlights the weighted behavioral codes for these populations, revealing distinctive patterns in their decision-making tendencies.

\end{document}